\documentclass[runningheads]{llncs}

\usepackage{eccv}

\usepackage{eccvabbrv}

\usepackage{graphicx}
\usepackage{booktabs}

\usepackage[accsupp]{axessibility}  % Improves PDF readability for those with disabilities.

\usepackage{hyperref}
\usepackage{makecell}

\usepackage{orcidlink}

\DeclareMathOperator*{\argmax}{arg\,max}

\usepackage{multirow}
\usepackage{booktabs}

\usepackage{tabularx}
\usepackage{adjustbox}
\usepackage{wrapfig}
\usepackage{amsmath}
\usepackage{wrapfig}
\usepackage{xspace}

\definecolor{analysisblue}{RGB}{31,119,180}
\definecolor{analysisorange}{RGB}{255,127,14}
\definecolor{analysispurple}{RGB}{123,104,238}

\newcommand{\app}[1]{Appendix~\ref{app:#1}}
\newcommand{\fig}[1]{Figure~\ref{fig:#1}}
\newcommand{\tab}[1]{Table~\ref{tab:#1}}
\newcommand{\sect}[1]{Section~\ref{sect:#1}}
\newcommand{\eq}[1]{Equation~\ref{eq:#1}}

\usepackage{hyperref}
\usepackage{xcolor}
\usepackage{mathtools}
\usepackage{subcaption}

\hypersetup{
    colorlinks=true,
    urlcolor=[rgb]{1,0.5,0.5}
}
\usepackage[table]{xcolor}

\begin{document}

% \title{Autoregressive Models Take the Lead in Generative Image Classification} 
\title{Revisiting Autoregressive Models for \\ Generative Image Classification}

\titlerunning{Revisiting AR Models for Generative Image Classification}

% \author{Ilia Sudakov \quad Artem Babenko  \quad Dmitry Baranchuk}
% \authorrunning{I.~Sudakov et al.}
% \institute{Yandex Research}

% \author{Ilia Sudakov\inst{2,1} \quad Artem Babenko\inst{1} \quad Dmitry Baranchuk\inst{1}}
% \authorrunning{I.~Sudakov et al.}
% \institute{\inst{1}Yandex Research \hspace{7mm} \inst{2}HSE University}

\author{Ilia Sudakov\inst{1,2} \quad Artem Babenko\inst{2} \quad Dmitry Baranchuk\inst{2}}
\authorrunning{I.~Sudakov et al.}
\institute{$^1$HSE University\hspace{7mm} $^2$Yandex Research}
% \institute{HSE University \and Yandex  Research}

\maketitle

\begin{abstract}
Class-conditional generative models have emerged as accurate and robust classifiers, with diffusion models demonstrating clear advantages over other visual generative paradigms, including autoregressive (AR) models. 
In this work, we revisit visual AR-based generative classifiers and identify an important limitation of prior approaches: their reliance on a fixed token order, which imposes a restrictive inductive bias for image understanding.
We observe that single-order predictions rely more on partial discriminative cues, while averaging over multiple token orders provides a more comprehensive signal. 
Based on this insight, we leverage recent any-order AR models to estimate order-marginalized predictions, unlocking the high classification potential of AR models.
Our approach consistently outperforms diffusion-based classifiers across diverse image classification benchmarks, while being up to $25{\times}$ more efficient.
Compared to state-of-the-art self-supervised discriminative models, our method delivers competitive classification performance -- a notable achievement for generative classifiers.
The code and models are available at: \url{https://github.com/yandex-research/ar-classifier}.

\end{abstract}

\section{Introduction}
\label{sec:intro}
Generative models (GMs) have recently shown a remarkable ability to approximate highly complex visual data distributions~\cite{peebles2023scalable, pang2025randar, tian2024visual, chang2022maskgit}.
This progress raises the question of whether they can also succeed in well-studied discriminative setups. % extend beyond generative tasks to
GMs have already demonstrated strong representation learning capabilities~\cite{kim2025revelio, yang2023diffusion, chen2024deconstructing, baranchuk2021label}, achieving comparable performance to state-of-the-art self-supervised discriminative approaches~\cite{chen2021empirical, oquab2023dinov2} on multiple downstream benchmarks.

Another line of research investigates whether GMs can be used directly as \textit{generative classifiers} (GCs).
Likelihood-based GMs, such as diffusion models (DMs) and autoregressive (AR) models, can approximate class-conditional likelihoods $p(\mathbf{x}|\mathbf{y})$ and then derive posterior predictions $p(\mathbf{y}|\mathbf{x})$ via Bayes’ rule.
Interestingly, recent works~\cite{li2024generative, jaini2023intriguing} reveal several intriguing properties of GCs. 
Unlike discriminative classifiers, which often rely on spurious correlations, GCs have been shown to avoid shortcut solutions~\cite{li2024generative}. 
Moreover, \cite{jaini2023intriguing} found that GCs exhibit shape-biased predictions that align more closely with human perception, whereas standard discriminative models are typically texture-biased.
% Interestingly, recent works~\cite{li2024generative, jaini2023intriguing} show the promise of modern GMs as GCs, revealing several intriguing properties. 

Motivated by the strong image generation quality of DMs, prior work~\cite{li2023your, li2024generative, jaini2023intriguing, li2025autoregressive} primarily focused on diffusion classifiers (DCs), leaving AR models relatively underexplored.
However, recent advances in visual AR model design~\cite{sun2024autoregressive, tian2024visual, pang2025randar, yu2025randomized, li2025fractal, wang2025parallelized, li2024autoregressive, luo2024open} have dramatically improved their generative performance, now achieving results competitive with DMs.
These developments motivate a timely re-examination of visual AR-based GCs. 

% In our work, we emphasize an important design choice of iterative visual GMs, such as DMs or AR models: \textit{how do they process images?}.
An important design choice affecting the performance of various iterative visual GMs is \textit{how they generate images}.
For example, classical next-token prediction AR models~\cite{sun2024autoregressive, yu2021vector} typically generate images using a \textit{raster-scan order}.
In contrast, DMs generate images in a coarse-to-fine manner, which can be interpreted as a form of \textit{spectral autoregression}~\cite{rissanen2023generative, dieleman2024spectral, yang2023diffusion1}.
From this perspective, DMs behave as implicit AR models that follow a low-to-high frequency order, better reflecting the hierarchical structure of natural images. 
We therefore hypothesize that this design choice is also critical for generative classification.

\begin{figure*}[t!]
% \vspace{3mm}
\centering
\includegraphics[width=0.98\linewidth]{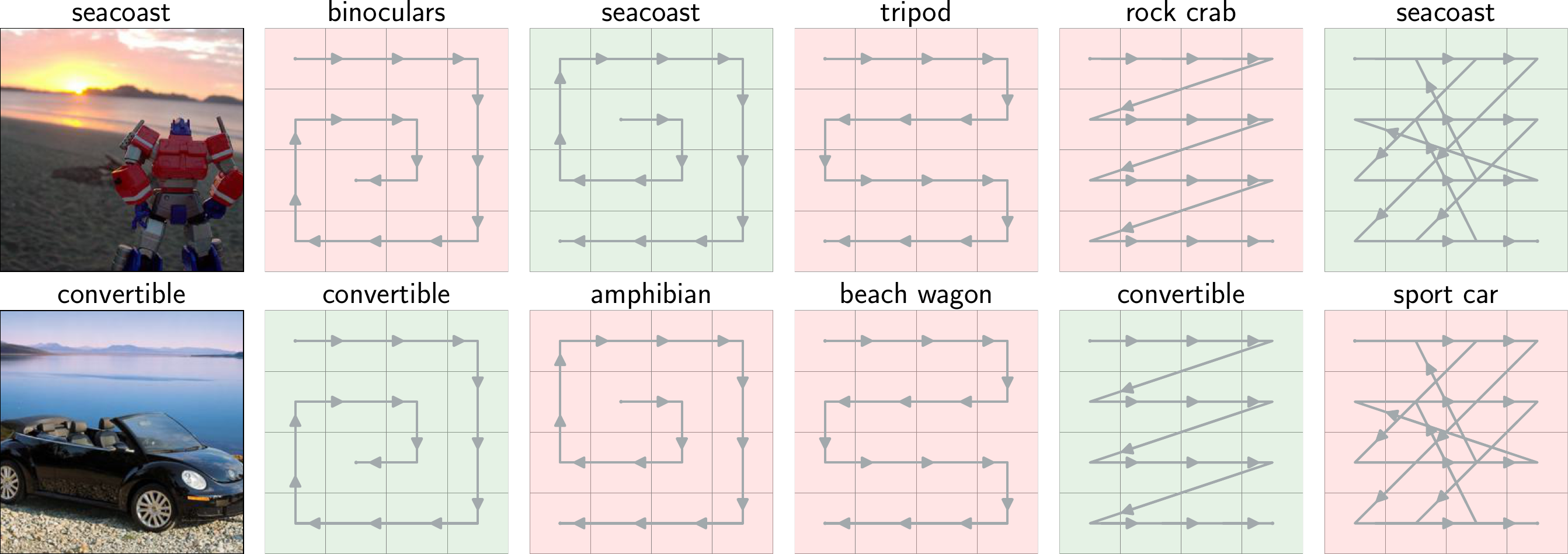}
\vspace{-1mm}
\caption{
     \textbf{Effect of different token orders on AR-based image 
     classification.} 
     Each row presents an image and
     five different token orders used by the any-order AR model to predict its class.
     Token orders can largely affect the final classification outcome.
}
\label{fig:motivation}
\end{figure*}
\paragraph{Contributions.}
In our work, we highlight the role of token ordering in AR-based generative classification.
In~\Cref{fig:motivation}, we provide an illustrative example by classifying several images using an AR model evaluated under different token orders.
We observe that some images are correctly classified under certain orders but misclassified under others, indicating that token orders may affect how the model interprets the image.
This observation is further analyzed in~\Cref{sect:random_vs_fixed}.

% To address this limitation, we investigate recent \textit{any-order} AR models that support generation under arbitrary token orders. 
% A practical advantage of these models is that they can evaluate \textit{order-conditional} log-likelihoods in a single forward pass, while DCs typically require $100{-}250$ model evaluations~\cite{li2023your}. 
% Although AR predictions for a single token order can be less accurate than those of DMs due to weaker visual inductive bias, any-order AR models can improve classification accuracy by marginalizing over multiple token orders. % while remaining an order of magnitude more efficient than DMs.
% Intuitively, averaging predictions across different orders may capture contextual information from multiple image regions, leading to more discriminative predictions.

To address this limitation, we investigate recent \textit{any-order} AR (AO-AR) models that support generation under arbitrary token orders. 
Although a single AR prediction for a fixed token order can be less accurate than that of DMs due to the weaker visual inductive bias, any-order AR models can improve classification accuracy by marginalizing over multiple token orders. % while remaining an order of magnitude more efficient than DMs.
Intuitively, averaging predictions across different orders may capture contextual information from multiple image regions, leading to more discriminative predictions.

A practical advantage of AR models is that they can evaluate \textit{order-conditional} log-likelihoods in a single forward pass, while DCs typically require $100{-}250$ model evaluations for a single likelihood estimate~\cite{li2023your}.
Therefore, even after marginalizing over multiple token orders, AR models remain an order of magnitude more efficient than DMs.

 %, an option that DMs do not provide.
% We note a practical advantage of the AR models of evaluating \textit{order-conditional} log-likelihoods in a single forward pass, while DCs typically require $100{-}250$ model evaluations~\cite{li2023your}. 
% Therefore, averaging over multiple orders appears a order of magnitude faster than a single DM evaluation.

Building on this insight, we propose an \textit{order-marginalized} AR-based generative classifier that achieves both higher accuracy and improved efficiency compared to diffusion-based classifiers. 
The experiments show that our AR-based approach outperforms diffusion classifiers in both in-domain accuracy and robustness to distribution shifts, while offering up to $25{\times}$ faster inference.

Furthermore, we compare GCs with state-of-the-art self-supervised discriminative models such as DINOv2~\cite{oquab2023dinov2}. 
We find that our AR-based approach offers competitive classification performance on both in-domain and out-of-distribution benchmarks. 
To our knowledge, this is the first work that compares GCs with state-of-the-art SSL approaches which are typically stronger baselines than standard supervised classifiers.

\section{Related work}
\label{sec:related}

\paragraph{Autoregressive models for image generation.}
Autoregressive (AR) models~\cite{sun2024autoregressive, yu2024language, esser2021taming, Yu2022Pathways, li2024autoregressive, pang2025randar} were a promising paradigm for image generation prior to the wide adoption of diffusion models (DMs).
They typically employ image tokenizers based on variants of VQ-VAE~\cite{van2017neural, chang2022maskgit, yu2024image, esser2021taming, yu2021vector} to map images into discrete token sequences.
Then, early AR models~\cite{esser2021taming, larochelle2011neural, van2016pixel, van2016conditional} performed \textit{next-token prediction}, generating images token by token, typically in a fixed raster order (left-to-right, top-to-bottom), similarly to language models~\cite{yenduri2023generative, radford2018improving, touvron2023llama}.
However, these models were later largely surpassed by DMs in generation quality, until more recent work introduced competitive next-token prediction~\cite{sun2024autoregressive, yue2025understand, yu2025randomized, pang2025randar, ren2025beyond} and parallel decoding~\cite{li2025autoregressive, tian2024visual, pang2025randar, wang2025parallelized, ren2025beyond} variants of AR models.

% Concurrently, other AR variants~\cite{li2025autoregressive, tian2024visual, pang2025randar, wang2025parallelized, ren2025beyond} have extended the next-token prediction idea to parallel decoding, i.e., predicting multiple tokens at a time.
% These models substantially accelerate inference while maintaining state-of-the-art image fidelity.

Importantly, most of these AR approaches fixate a specific token order. % for the model.
As we show later, this detail imposes an important limitation for generative classification with AR models.
Therefore, in our work, we focus on one of the recent next-token prediction models, RandAR~\cite{pang2025randar}, which, unlike most AR counterparts, explicitly conditions on token positions and supports generation in arbitrary token orders.
We will discuss this model in more detail in~\sect{background}.

% While diffusion models (DMs) have long dominated in the field, recent AR models could reach competitive performance~\cite{lammagen, var, randar, rar, arpg, xar, par}.
% % The AR models can be grouped in two major categories: \textit{next-token prediction} and \textit{parallel token prediction}.
% Next-token prediction models~\cite{lammagen, rar, randar} generate images token-by-token, similar to LLMs.
% RAR~\cite{} and RandAR~\cite{} has shown that these models can benefit from training on random token orders. 
% Another research direction explored parallel-decoding enabling more efficient sampling without compromising the generative performance.
% VAR~\cite{} proposed a more natural AR modeling for images by progressively generating images scale-by-scale rather than using next-token prediction.
% PAR~\cite{} and xAR~\cite{} consider various strategies.
% \TODO{}.

\paragraph{Generative classifiers.}

% On the other hand, GCs have a significant limitation in terms of efficiency since models need to be applied to each class. 

Among diverse generative paradigms~\cite{ho2020denoising, song2020score, lai2025principles, lai2025principles, esser2021taming, kingma2013auto, goodfellow2020generative}, diffusion classifiers (DCs)~\cite{li2023your,li2024generative, chen2024diffusion, chen2023robust} have recently received increased attention.
DCs achieve competitive in-domain performance with discriminative models while offering improved robustness under distribution shifts~\cite{li2024generative}.
However, an important limitation of DCs is their notoriously slow inference process that typically requires hundreds of model forward passes per image to achieve high accuracy.
Joint Energy-based Models (JEMs)~\cite{guo2023egc, grathwohl2019your, yin2025joint} address this problem by combining discriminative and generative objectives within a unified framework.

Meanwhile, classical AR models have proven effective as GCs in text classification tasks~\cite{li2024generative, kasa2025generative}, though they were outperformed by DCs in visual domains~\cite{jaini2023intriguing}. 
Recent work~\cite{chen2025your} explores the potential of modern visual AR models by introducing VAR-based GCs with contrastive alignment fine-tuning.
In our work, we investigate GCs built upon state-of-the-art next-token prediction models and focus on purely generative approaches, without any additional tuning.

% \todo{Hybrid approaches such as DAT~\cite{yin2025joint} seek to bridge this gap by combining discriminative and generative objectives within a unified framework.} 

% While previously explored tuning-free DCs were , the recent work~\cite{noise_matters} showed that the additional noise optimization may significantly improve the DC performance. 

% Nevertheless, the recent work~\cite{noise_matters} showed that optimizing the noise for DC may handle category-related frequency and spatial information and improve its performance even further.

% With a drastic development of generative models, the question if they can be effective for discriminative problems appears more clearly.
% Multiple works have explored the GMs as feature extractors and showed that recent DMs are effective unsupervide learners, producing highly informative features~\cite{label, dae} and compete with state-of-the-art discriminative self-supervised approaches~\cite{moco, dino} in downstream tasks.
% % 

\section{Preliminaries}
\label{sect:background}
% DONE

\paragraph{Generative Classifiers.}
In general, GCs rely on estimating class-conditional likelihoods $p(\mathbf{x} | c)$ and apply Bayes’ rule to obtain posterior probabilities $p(c | \mathbf{x}) \propto p(\mathbf{x}|c) p(c)$.
Then, classification can be performed as: 
\begin{equation}
    \label{eq:main}
    c_i = \argmax_{c_i \in C} \left[ \log p(\mathbf{x}|c_i) + \log p(c_i) \right],
\end{equation}
where the $\log p(c_i)$ term is often ignored in practice, assuming a uniform prior $p(c)$ over classes.
% A uniform prior $p(c)$ over $\{c_i\}$ (i.e., $p(c_i) = \frac{1}{N}$) is natural and
% leads to all of the $\log p(c)$ canceling. 

Pretrained DMs $\epsilon_\theta$ approximate $\log p(\mathbf{x}|c)$ by estimating ELBO: 
\begin{equation}
    \label{eq:elbo}
    \log p(\mathbf{x}|c) \geq ELBO(\mathbf{x}, c) \approx \sum_{t, \epsilon} \lambda_t \Vert\epsilon_\theta(\mathbf{x}_t, t, c) - \epsilon \Vert^2,
    \vspace{-2mm}
\end{equation}
where $\mathbf{x}_t$ is obtained using the forward diffusion process $q(\mathbf{x}_t | \mathbf{x}) = \mathcal{N}(\mathbf{x}_t | \alpha_t \mathbf{x}, \sigma^2_t \mathbf{I})$.
Previous works~\cite{li2023your, li2024generative} observed that using an objective with uniform weighting $\lambda_t=1$ yields better classification performance in practice.
Note that, \Cref{eq:elbo} requires $100{-}250$ model evaluations across different timesteps~\cite{li2023your}.
Since most state-of-the-art DMs are latent diffusion models~\cite{Rombach_2022_CVPR}, classification accuracy is computed in the continuous VAE latent space.

In contrast, AR models first encode images into discrete token sequences $\mathbf{x}=\{\mathbf{x}_1, ..., \mathbf{x}_N\}$ of length $N$, typically using a VQ-VAE.
Then, they provide explicit log-likelihood estimates in the latent space using a single forward pass:
\begin{equation}
    \label{eq:ar_likelihood}
    \log p(\mathbf{x}|c) \approx \sum^N_{n=1} \log p_\theta(\mathbf{x}_n | \mathbf{x}_{<n}, c)
\end{equation}

In our work, we highlight that both DMs and AR models implicitly impose an inductive bias through the way images are modeled, and that this choice can directly affect classification performance.

\paragraph{Any-order Autoregressive Modeling.}
% Suggesting that raster-order may be suboptimal and unnatural choice for visual modeling, recent works have explored more promising alternatives such as scale-wise modeling~\cite{} and random-order models~\cite{}. 

In our work, we aim to explore the role of different AR token orders on image classification and therefore consider RandAR~\cite{pang2025randar}, a state-of-the-art AR model capable of generating images in arbitrary token orders.
RandAR augments the image token sequence $\mathbf{x}=\{\mathbf{x}_1, ..., \mathbf{x}_N\}$ with a corresponding set of \textit{position instruction tokens} $P=\{\mathbf{p}_1, ... \mathbf{p}_N\}$.
Each position token $\mathbf{p}_i$ is paired with its respective image token, forming an interleaved sequence $[\mathbf{p}_1, \mathbf{x}_1, ..., \mathbf{p}_N, \mathbf{x}_N]$. 

To enable any-order generation, RandAR applies a random permutation $\pi$ over token indices, producing the reordered sequence $[\mathbf{p}^{\pi(1)}_1, \mathbf{x}^{\pi(1)}_1, ...,\mathbf{p}^{\pi(N)}_{N}, \mathbf{x}^{\pi(N)}_N$], where $\pi(i)$ denotes the original position of the $i$-th token.
Thus, RandAR \textit{explicitly} defines the order-conditional likelihood $p(\mathbf{x}|\pi, c)$ as
\begin{equation}
p(\mathbf{x}|\pi, c) = \prod^N_{n=1} p(\mathbf{x}^{\pi(n)}_n | \mathbf{p}^{\pi(1)}_1, \mathbf{x}^{\pi(1)}_1, ..., \mathbf{x}^{\pi(n-1)}_{n-1}, \mathbf{p}^{\pi(n)}_n, c)
\label{eq:randar}
\end{equation}
% \mathbb{E_{\pi}} 
Note that RandAR can be readily reduced to a raster-order AR model by fixing the permutation to the identity mapping $\pi(i)=i$.

% \todo{RandAR is not the only architecture that supports random-order generation, but it provides public checkpoints suitable for evaluation.
% Another token-order-conditioned model is RAR~\cite{yu2025randomized}. However, it is not natively any-order, as it is trained with random permutations that are gradually annealed to the raster order.
% In~\sect{masked_diffusion}, we additionally adapt RAR for any-order sampling and show that our order-marginalized approach is not specific to RandAR.
% }

% In the following, we will use the RandAR architecture all over our analysis to conduct fair comparisons within the  
% same architecture and training setups.
% Note that RandAR can be readily reduced to the raster-order AR model by fixing the identity permutation.

\begin{figure*}[t!]
% \vspace{-1mm}
\centering
\includegraphics[width=\linewidth]{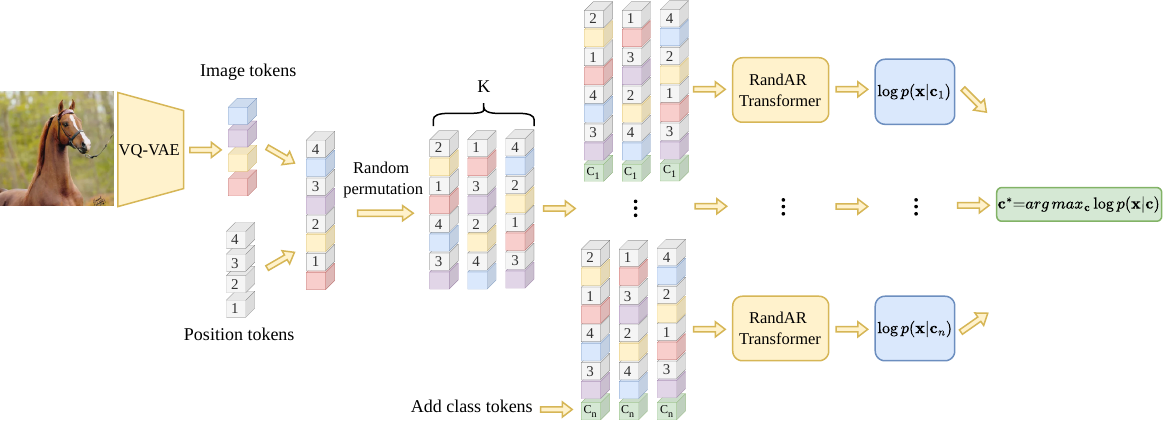}
\vspace{-5mm}
\caption{
    \textbf{Order-marginalized generative classification framework.} 
    i) Input image is tokenized with the VQ-VAE into a sequence of discrete tokens.
    ii) Position-instruction tokens are concatenated with the image tokens, and $K$ randomly permuted sequences are constructed.
    iii) Class condition tokens $c_i$ for each class are appended to every sequence.
    iv) The RandAR model predicts $\log p(\mathbf{x}|c_i)$ using \eq{randar_likelihood}.
    v) The predicted class $c^*$ is obtained as $\argmax_{c_i} \log p(\mathbf{x}|c_i)$.
}
\label{fig:main_scheme}
\end{figure*}

\section{Order-marginalized generative classification}
\label{sect:method}
% DONE
 % rather than a single $p(\mathbf{x}|c)$ used in~\eq{main}.

Since RandAR explicitly conditions on token positions and models multiple $p(\mathbf{x}|\pi,c)$, it allows estimating the order-unconditional likelihood $p(\mathbf{x}|c)$, by marginalizing over all possible permutations as $\mathop{\mathbb{E}}_{\pi} \left[ p(\mathbf{x} | \pi,c) \right]$.
Then, this expectation can be estimated with Monte Carlo sampling using $K$ random order samples.
However, in practice, we find this estimate of $p(\mathbf{x}|c)$ significantly less effective than estimating the lower bound on the order-unconditional log-likelihood, derived using Jensen’s inequality:
\begin{equation}
    \log p(\mathbf{x} | c) = \log \mathop{\mathbb{E}}_{\pi} \left[ p(\mathbf{x} | \pi, c) \right] \geq \mathop{\mathbb{E}}_{\pi} \left[ \log p(\mathbf{x} | \pi, c) \right] \approx \frac{1}{K} \sum^K_{k=1} \log p(\mathbf{x} | \pi_k, c) 
\label{eq:randar_likelihood}
\end{equation}

For classification, we then append class tokens $C=\{\mathbf{c}_1,...,\mathbf{c}_M\}$ to each permuted sequence, producing $[M, K]$ input sequences of length $2N + 1$.
Importantly, the same sampled orders are shared across all classes to ensure consistency.

For each class token $\mathbf{c}_i$, the model produces $[\log p(\mathbf{x} | \pi_1, \mathbf{c}_i), ..., \log p(\mathbf{x} | \pi_K, \mathbf{c}_i)]$, which are then aggregated into $\log p(\mathbf{x} | \mathbf{c}_i)$ using~\eq{randar_likelihood}.
Finally, the class $c^*$ is predicted according to~\eq{main}.
The overall scheme of our classification procedure is presented in~\fig{main_scheme}.

\paragraph{Discussion.}
We hypothesize that the lower bound largely outperforms the direct likelihood estimate because RandAR was trained directly on~\eq{randar_likelihood}, making it better aligned with image understanding. 
Notably, classification with DMs also gets worse when evaluated using the true ELBO estimate, and instead show significantly better classification results when evaluated with their training objective~\Cref{eq:elbo} with $\lambda_{t}=1$~\cite{li2023your}.
% We confirm this observation in our ablation study in \sec{}.

Another question is whether we can similarly marginalize DM predictions.
According to~\Cref{eq:elbo}, DMs allow marginalization over multiple noise samples at each timestep.
However, in \app{average_dm}, we show that additional averaging over noise samples provides no noticeable improvements and only increases computational cost.
We hypothesize that this is because noise averaging is already included during timestep averaging. 
% and additional averaging per timestep yields only marginal gains. 
% Instead, we observe that DMs benefit primarily from averaging across timesteps, which we carefully account for in our experiments.

%change how DMs process images as they still follow the same spectral ordering~\cite{dieleman2024spectral}.

Finally, AO-AR models are closely connected to \textit{Masked Diffusion} (MD) models~\cite{ou2024absorbing, hoogeboom2022autoregressive}.
In~\app{ao_ar_md}, we show that the order-marginalized AO-AR objective in~\Cref{eq:randar_likelihood} equals the MD evidence lower bound in expectation.
This connection enables both AR- and MD-based evaluation of any-order AR-based GCs.
In Section~\ref{sect:masked_diffusion}, we show that the AR formulation consistently outperforms the MD one.

\begin{figure*}[t!]
\centering
\includegraphics[width=0.75\linewidth]{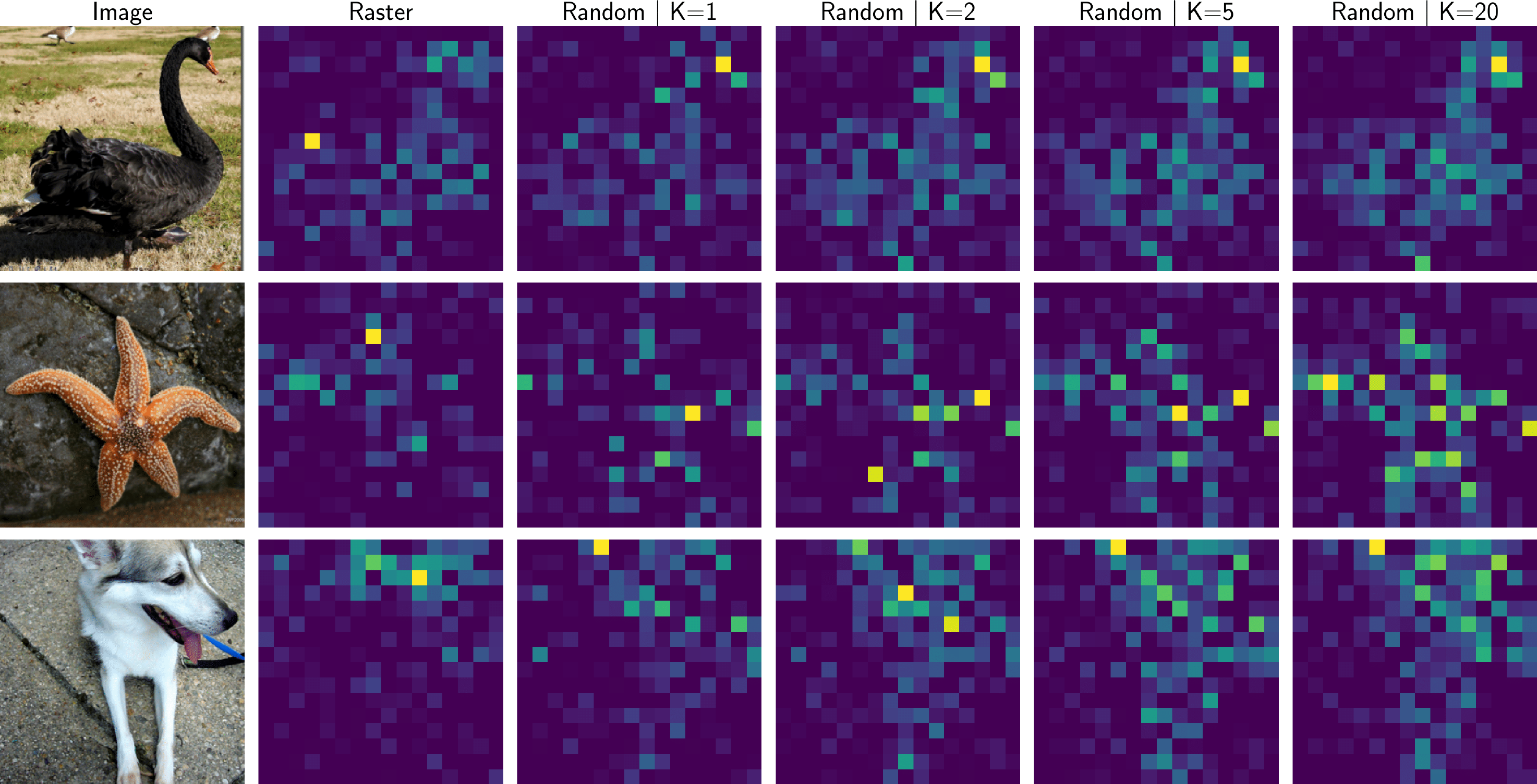}
\vspace{-1mm}
\caption{
    % DONE
    Per-token ``discriminative'' log-likelihoods, computed as $clip[\log p(\mathbf{x} | c_{true}) - \log p(\mathbf{x} | c_{false}), 0]$, across different orders and $K$ values.
    $c_{true}$ denotes the correct class; $c_{false}$ refers to a randomly selected incorrect one.
    Order-marginalized log-likelihood estimates ($K>1$) capture class-specific attributes more accurately.
}
\label{fig:per_token_likelihoods}
\end{figure*}

\section{Analysing order-marginalized AR classifiers}
% \section{Order-marginalized vs fixed-order AR classifiers} 
\label{sect:random_vs_fixed}
% In this section, we explore the effect of order-marginalized log-likelihood estimation on image classification.

\paragraph{Setup.}
% DONE
We pretrain two RandAR-L/16 variants under same training setups following~\cite{pang2025randar}: a model using a fixed raster-order and another operating with random orders.
The models are trained on ImageNet1K~\cite{russakovsky2015imagenet} using $256{\times}256$ images.
We use the LlamaGen tokenizer~\cite{sun2024autoregressive}, producing sequences of $N{=}256$ tokens.
% We use LammaGen~\cite{} VQ-VAE for image size of $256{\times}256$ resulting in $N=256$ tokens in a sequence.

% \subsection{Randomized vs fixed-order AR classifiers}
% \label{sect:random_vs_fixed}
% First, we explore how AR-based classifiers perform with different ordering strategies.
% Intuitively, random-order may provide more discriminative per-token likelihoods due to capturing global image context at each step.

\paragraph{Per-token likelihoods.}
% DONE
First, we estimate per-token ``discriminative'' log-likelihoods  $clip[\log p(\mathbf{x} | c_{true}) - \log p(\mathbf{x} | c_{false}), 0]$, where $c_{true}$ denotes the correct class and $c_{false}$ a random incorrect one.
These log-likelihoods indicate which tokens are more discriminative for a particular class prediction.

\fig{per_token_likelihoods} shows the results for several images, with more examples in \app{per_token_likelihoods}.
We observe that the raster-order model focuses more on specific class-object parts.
Similarly, the random-order variant with $K{=}1$, a single-sample Monte Carlo estimate of \eq{randar_likelihood}, also focuses on localized object parts.
In contrast, for $K{=}20$, the object shapes appear more clearly, indicating that the model leverages broader class-level information for classification. 

\begin{figure*}[t!]
\centering
\includegraphics[width=0.87\linewidth]{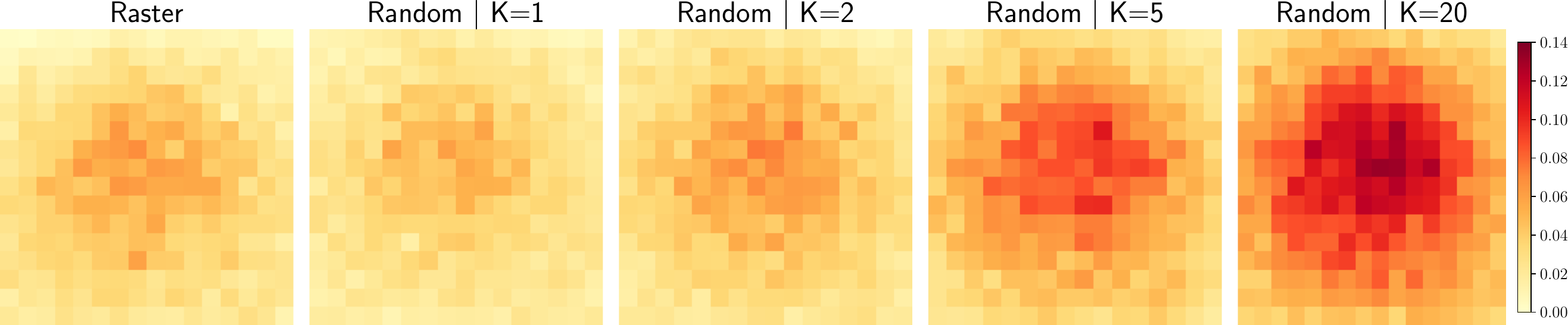}
\vspace{-1mm}
\caption{
    %DONE
    Per-token accuracy of the RandAR classifier across different orders and $K$ values. 
    Center image tokens appear more discriminative, likely due to the center-object bias present in ImageNet.
    Increasing $K$ consistently improves accuracy for all tokens.
}
\label{fig:per_token_accuracy}
\end{figure*}

\paragraph{Per-token accuracy.}
% DONE
\fig{per_token_accuracy} presents per-token top-1 accuracy averaged over $1000$ images, shown at their original token positions.
The token accuracy consistently increases with $K$, resulting in more discriminative tokens. 
Central image tokens achieve higher accuracy, reflecting ImageNet’s center-object bias.

% \begin{table}[t!]
% % \centering

% \resizebox{0.5\linewidth}{!}{
%     \begin{tabular}{l|c|cccc}
%     \toprule
%     Token order & IN-Val & IN-R & IN-S & IN-A \\
%     \midrule
%      Raster      & $0.701$ & $0.351$ & $0.301$ & $0.174$\\
%      Random $K{=}1$  & $0.670$ & $0.351$ & $0.289$ & $0.110$\\
%      \midrule
%      Random $K{=}2$  & $0.726$ & $0.402$ & $0.343$ & $0.127$\\
%      Random $K{=}5$  & $0.759$ & $0.444$ & $0.383$ & $0.140$\\
%      Random $K{=}10$ & $0.768$ & $0.459$ & $0.392$ & $0.144$ \\
%      Random $K{=}20$ & $\mathbf{0.769}$ & $\mathbf{0.469}$ & $\mathbf{0.409}$ & $\mathbf{0.146}$ \\
%     \bottomrule
%     \end{tabular}
% }
% \caption{Comparison of RandAR trained with random and raster orders in terms of top-1 classification accuracy on ImageNet1K.}
% \label{tab:imagenet_raster_vs_random}
% \end{table}

\begin{figure}[t]
\vspace{10pt}
\begin{minipage}{0.52\linewidth}
\includegraphics[width=\linewidth]{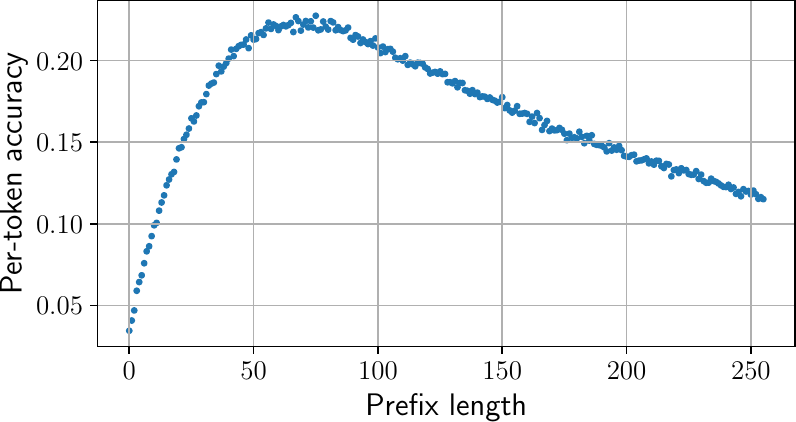}
\vspace{-5mm}
\caption{
    Per-token accuracy of the RandAR classifier for $K{=20}$ w.r.t. the prefix length. 
}
\label{fig:per_token_accuracy_by_prefix}
\end{minipage} \hfill
\begin{minipage}{0.44\linewidth}

\resizebox{\linewidth}{!}{
\begin{tabular}{l|c|cccc}
\toprule
Token order & IN-Val & IN-R & IN-S & IN-A \\
\midrule
 Raster      & $0.701$ & $0.351$ & $0.301$ & $0.174$\\
 Random $K{=}1$  & $0.670$ & $0.351$ & $0.289$ & $0.110$\\
 \midrule
 Random $K{=}2$  & $0.726$ & $0.402$ & $0.343$ & $0.127$\\
 Random $K{=}5$  & $0.759$ & $0.444$ & $0.383$ & $0.140$\\
 Random $K{=}10$ & $0.768$ & $0.459$ & $0.392$ & $0.144$ \\
 Random $K{=}20$ & $\mathbf{0.769}$ & $\mathbf{0.469}$ & $\mathbf{0.409}$ & $\mathbf{0.146}$ \\
\bottomrule
\end{tabular}}
\vspace{-3mm}
\captionof{table}{Comparison of RandAR trained with random and raster orders in terms of top-1 classification accuracy on ImageNet1K.}
\label{tab:imagenet_raster_vs_random}
\end{minipage}
% \vspace{-3mm}
\end{figure}

Also, we explore how token accuracy varies with prefix length.
\fig{per_token_accuracy_by_prefix} reports per-token accuracy for $K{=}20$, averaged over $30,000$ images.
Token accuracy initially increases with prefix length, peaking ${\sim}70$ tokens, after which it declines.
We hypothesize that the tokens with $[50-80]$ prefix lengths are more discriminative because such prefixes capture only high-level image information, forcing the model to generate class-defining details. In contrast, later tokens may contribute less, as they can rely more heavily on nearby observable tokens.
% \TODO{We further develop this intuition in~\app{attention_maps}.}

% Interestingly, although the likelihood of the i-th token is estimated at $K{=}20$ different random positions, the overall trend is similar to the pattern previously observed in raster-order AR models~\cite{yue2025understand}.

\paragraph{Overall performance.}
% DONE
In~\tab{imagenet_raster_vs_random}, we compare the models in terms of their overall classification performance on the ImageNet validation set and several out-of-distribution (OOD) benchmarks. 

We notice that the raster-order variant provides better results compared to the random-order one with $K{=}1$.
We attribute this to the limited model capability to fit arbitrary orders similarly well as a single one.
This intuition is supported by slightly worse generative performance of random-order models compared to the raster-order counterpart~\cite{pang2025randar}.
However, for $K{>}1$, the random-order model significantly outperforms the raster one across all validation sets.

Note that the higher accuracy of the random-order model for $K{>}1$ comes with additional computational cost, as each order requires an extra model forward pass.
Nevertheless, as shown in~\sect{efficiency}, this overhead is fully justified compared to diffusion-based classifiers, which typically require hundreds of model forward passes~\cite{li2024generative, li2023your}.

\section{Experiments}
\label{sec:exps}
% \TODO{add NFEs to the table.}
In this section, we compare order-marginalized AR classifiers with state-of-the-art discriminative and generative approaches.

\subsection{Main results}
\label{sect:main_results}

\paragraph{Baselines.}
% DONE
We evaluate several baseline families known for their strong classification performance:
\begin{itemize}
    \item \textbf{Discriminative approaches.} 
    We compare with supervised pretrained ViT~\cite{dosovitskiy2020image} models and the state-of-the-art self-supervised learning (SSL) method DINOv2~\cite{oquab2023dinov2} with linear probing.
    To our knowledge, prior works have not compared against leading discriminative SSL methods, despite their largely superior performance over standard supervised classifiers.
    
    \item \textbf{Diffusion classifiers.} 
    For DM-based baselines, we consider DiT~\cite{peebles2023scalable} and SiT~\cite{ma2024sit}.  
    Since SiT is based on flow matching~\cite{lipman2022flow}, we observed that classification using the $v$-prediction led to suboptimal results.
    Therefore, we estimate ELBO for SiT using the $\epsilon$-prediction objective, which consistently yields better performance.
    In~\app{loss_weighting}, we ablate various ELBO loss weighting schemes for both DiT and SiT to ensure a fair comparison.
    Also, we evaluate SiT equipped with representation alignment (REPA)~\cite{yu2024representation}, which leverages SSL representations to improve generative performance.
    In~\app{few-step}, we further compare against MeanFlow~\cite{geng2026mean}, a recent few-step DM-based model.

    \item \textbf{Joint generative-discriminative models.} 
    We also compare to the most recent state-of-the-art joint energy-based model, DAT~\cite{yin2025joint}.
    \item \textbf{Autoregressive models.} 
    We compare our method with LlamaGen~\cite{sun2024autoregressive}, VAR~\cite{tian2024visual} and A-VARC+~\cite{chen2025your}.
    For VAR, likelihoods are estimated across all spatial scales. 
    A-VARC+ additionally employs likelihood smoothing for more accurate classification.
    Also, we pretrain RandAR with a fixed raster-order to take into account the model design differences with LlamaGen. 
\end{itemize}

All baselines are pretrained on ImageNet1K~\cite{russakovsky2015imagenet}.
For methods lacking corresponding publicly available checkpoints, we reproduced the pretraining ourselves, strictly following their official implementations.
We provide more details in~\app{setup}.

To ensure a fair comparison, all models share similar transformer-based architectures, comparable model sizes, same image resolution $256{\times}256$ and the number of tokens per image ($256$), thereby minimizing the factors unrelated to model types.
The only exception is DAT, which uses a ConvNext-L backbone.

\paragraph{Datasets.}
% DONE
For evaluation, we use the ImageNet-Val set and several OOD benchmarks such as ImageNet-R/S/A~\cite{hendrycks2021natural, wang2019learning, hendrycks2021many} and ImageNet-C~\cite{hendrycks2019benchmarking} for Gaussian noise and JPEG corruptions applied with a strength 3 out of 5.
For all validation sets, we report top-1 accuracy.
Since GCs require substantial compute for $1000$-class predictions, we follow the protocol of~\cite{li2023your} and report results on 2K subsets of IN-Val, IN-S and IN-C.
We verified that increasing the subset size yields only minor variations in accuracy.
$200$-class benchmarks, IN-R and IN-A, are evaluated on the entire sets.

\begin{table*}[t!]
\centering
\begin{adjustbox}{width=\textwidth}
\begin{tabular}{c|c|l|cccccc}
\toprule
Model size & Model type & Method & IN-Val & IN-R & IN-S & IN-C Gauss & IN-C JPEG & IN-A \\
\midrule
\multirow{9}{*}{L/16} 
 & \multirow{2}{*}{Discriminative} & ViT~\cite{dosovitskiy2020image}      & $0.803$ & $0.409$ & $0.291$ & $\textbf{0.620}$ & $0.694$ & $0.166$ \\
 & & DINOv2~\cite{oquab2023dinov2}   & $\textbf{0.819}$ & $\textbf{0.476}$ & $0.358$ & $0.497$ & $\textbf{0.736}$ & $\textbf{0.363}$ \\
 \cmidrule(lr){2-9}
& Joint model & DAT~\cite{yin2025joint}      & $0.764$ & $0.379$ & $0.364$ & $0.569$ & $0.709$ & $0.034$ \\
 \cmidrule(lr){2-9}
 & Diffusion & DiT~\cite{peebles2023scalable}      & $0.771$ & $0.393$ & $0.361$ & $0.467$ & $0.642$ & $0.133$ \\
\cmidrule(lr){2-9}
 & \multirow{5}{*}{Autoregressive} & LlamaGen~\cite{sun2024autoregressive} & $0.640$ & $0.298$ & $0.232$ & $0.358$ & $0.452$ & $0.143$ \\
 & & VAR~\cite{tian2024visual}      & $0.656$ & $0.255$ & $0.177$ & $0.137$ & $0.390$ & $0.083$ \\
 % & & A-VARC~\cite{chen2025your} & $0.683$ & $0.233$ & $0.157$ & $0.155$ & $0.377$ & $0.082$ \\
 & & A-VARC+~\cite{chen2025your} & $0.717$ & $0.277$ & $0.175$ & $0.161$ & $0.391$ & $0.072$ \\
 & & RandAR raster~\cite{pang2025randar} & $0.701$ & $0.351$ & $0.301$ & $0.476$ & $0.588$ & $0.174$ \\
 & & RandAR~\cite{pang2025randar} \textbf{(Ours)} & $0.780$ & $0.463$ & $\textbf{0.406}$ & $0.589$ & $0.687$ & $0.145$ \\
\midrule
\midrule
\multirow{8}{*}{XL/16}
 & Discriminative & DINOv2~\cite{oquab2023dinov2}   & $\textbf{0.827}$ & $0.486$ & $0.354$ & $0.500$ & $\textbf{0.754}$ & $\textbf{0.345}$ \\
\cmidrule(lr){2-9}
 & \multirow{3}{*}{Diffusion} & SiT~\cite{ma2024sit}      & $0.697$ & $0.257$ & $0.223$ & $0.311$ & $0.537$ & $0.104$ \\
 & & SiT + REPA~\cite{yu2024representation}      & $0.733$ & $0.296$ & $0.262$ & $0.416$ & $0.554$ & $0.169$ \\
 & & DiT~\cite{peebles2023scalable}      & $0.772$ & $0.402$ & $0.367$ & $0.505$ & $0.661$ & $0.153$ \\
\cmidrule(lr){2-9}
 & \multirow{4}{*}{Autoregressive} & LlamaGen~\cite{sun2024autoregressive}   & $0.559$ & $0.248$ & $0.200$ & $0.359$ & $0.391$ & $0.210$ \\
 & & VAR~\cite{tian2024visual}      & $0.630$ & $0.222$ & $0.153$ & $0.146$ & $0.336$ & $0.090$ \\
 % & & A-VARC~\cite{chen2025your} & $0.661$ & $0.199$ & $0.138$ & $0.077$ & $0.315$ & $0.090$ \\
 & & A-VARC+~\cite{chen2025your} & $0 .719$ & $0.259$ & $0.143$ & $0.051$ & $0.353$ & $0.083$ \\
 & & RandAR~\cite{pang2025randar} \textbf{(Ours)} & $0.813$ & $\textbf{0.530}$ & $\textbf{0.459}$ & $\textbf{0.652}$ & $\textbf{0.751}$ & $0.233$ \\
\bottomrule
\end{tabular}
\end{adjustbox}
\vspace{1mm}
\caption{Top-1 accuracy on various ImageNet benchmarks for different model types.}
\label{tab:imagenet_results}
\end{table*}

% \begin{wrapfigure}{l}{0.4\textwidth}
% \vspace{-8mm}
% \centering
% \includegraphics[width=\linewidth]{figures/per_token_accuracy_by_prefix_30k.pdf}
% \vspace{-7mm}
% \caption{
%     Per-token accuracy of the RandAR classifier for $K{=20}$ w.r.t. the prefix length. 
% }
% \vspace{-5mm}
% \label{fig:per_token_accuracy_by_prefix}
% \end{wrapfigure}

% \begin{wrapfigure}{l}{0.48\textwidth}
%     \centering
%     \includegraphics[width=\linewidth]{figures/dinov2_vs_randar_uncertainty.pdf}
%     \caption{
%     Comparison of uncertainty estimates between the RandAR and DINOv2 models for correctly and incorrectly predicted labels on IN-Val. 
%     RandAR provides more reliable uncertainty estimates compared to DINOv2.
% }
%     \label{fig:uncertainty}
% \end{wrapfigure}

\paragraph{Evaluation setup.}
% DONE
For diffusion-based classifiers and our RandAR-based ones, we provide the results using sufficiently large $K$ for RandAR and the number of timesteps for DiT/SiT to ensure near-optimal performance.
Specifically, we use $K{=}20$ and $250$ timesteps, following the setup in~\cite{li2023your}.
The RandAR models are trained with the LlamaGen tokenizer~\cite{sun2024autoregressive} with latent noise augmentation, which is discussed and ablated in~\sect{vqvae_analysis}. % and we explore the effect of image tokenizers on classification performance

\paragraph{Results.}
The results are presented in~\tab{imagenet_results}.
First, RandAR demonstrates strong scaling potential, showing significant performance gains as model size increases from L to XL across all validation sets.
Also, RandAR significantly surpasses fixed-order counterparts such as LlamaGen, VAR and the raster-order RandAR variant. 
Moreover, RandAR outperforms recent A-VARC, which averages predictions over multiple noise samples, resulting in comparable compute.

Compared to diffusion-based classifiers, RandAR achieves significantly superior performance on each validation set, with more pronounced gains on the OOD benchmarks.
Although it is beyond the scope of this work, we would like to note that SiT, while underperforming DiT, may benefit from future exploration of alternative weighting functions and timestep schedules.

% Comparing to the supervised classifier, RandAR achieves comparable in-domain accuracy while significantly  outperforming them on IN-R and IN-S benchmarks.
% Among the baselines, DINOv2 with linear probing appears the strongest competitor, exceeding RandAR’s in-domain accuracy by $2.5\%$ for the XL model.
Comparing to DINOv2, RandAR shows lower in-domain accuracy by $1.4\%$ for the XL model. 
Nevertheless, RandAR-XL surpasses DINOv2 on 3 of 5 OOD sets (IN-R, IN-S, IN-C Gauss), matches performance on IN-C JPEG, and only trails on IN-A.

% \todo{Comparing to hybrid models, RandAR outperforms DAT on all datasets except JPEG, where it has lower accuracy by $2.1\%$.}

Overall, RandAR establishes a new state-of-the-art among GCs and competes with DINOv2 — the result that has not been previously achieved with GCs. 

% \paragraph{DINOv2 linear probing vs kNN zero-shot classifier.}
% Also, we explore if linear probing may reduce the model's robustness to distribution shifts.
% To this end, we compare the linear probing setting with zero-shot kNN classifier on top of DINOv2 features across different validation sets.

\begin{figure*}[t!]
% \vspace{2mm}
\centering
\includegraphics[width=0.49\linewidth]{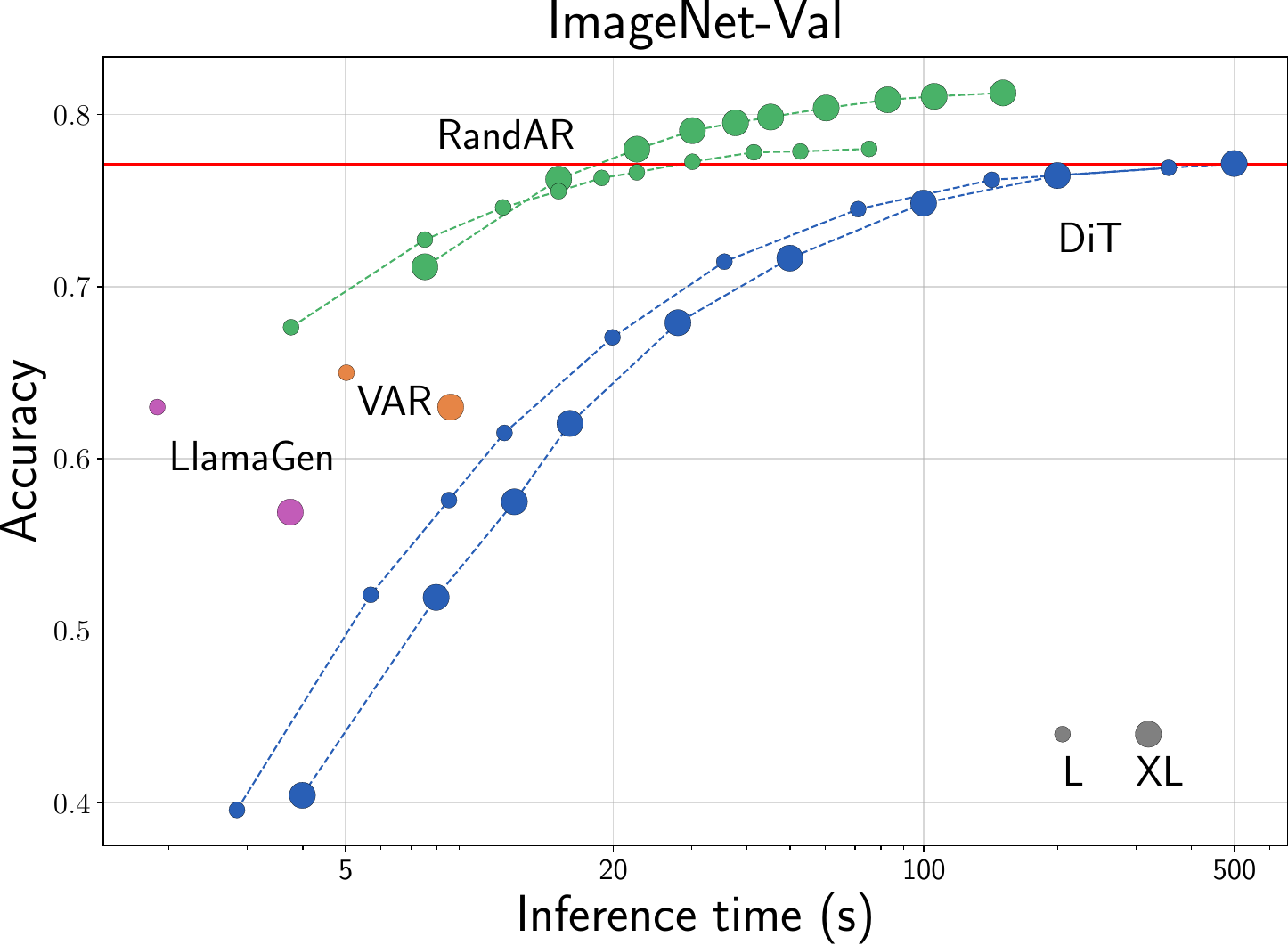}
\hfill
\includegraphics[width=0.49\linewidth]{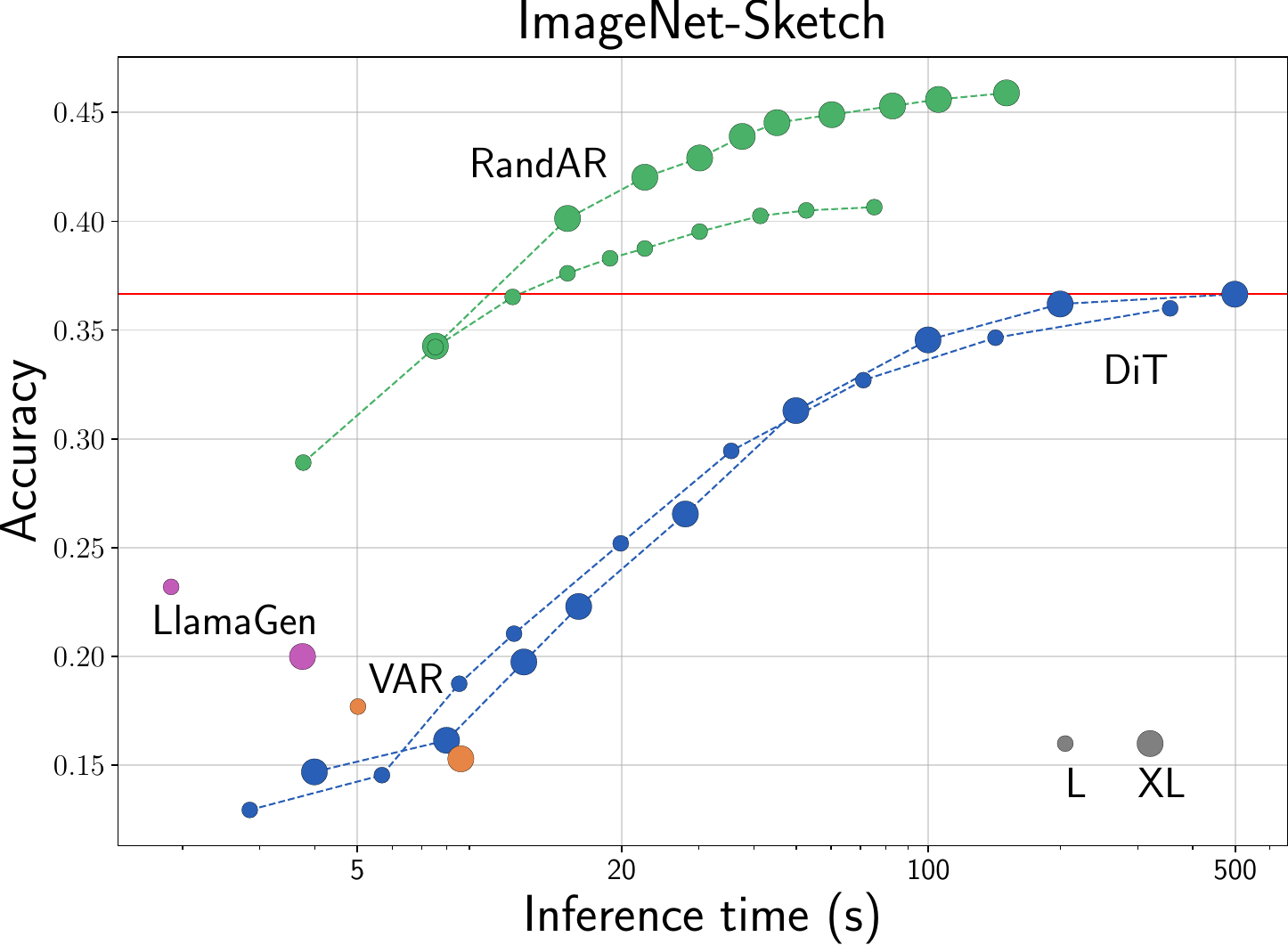}
% \vspace{-2mm}
\caption{
    Comparison of the generative classifiers in terms of top-1 classification accuracy and efficiency across different model sizes.
    Efficiency is reported in seconds per single-image classification.
}

\label{fig:efficiency}
\end{figure*}

\subsection{Efficiency evaluation}
\label{sect:efficiency}
% DONE
% Here, we evaluate various generative classifiers of different model sizes in terms of accuracy w.r.t. runtime in seconds per a single image classification.
% RandAR and DiT classifiers are evaluated with different NFEs to approximate the respective objectives, $p(\mathbf{x}|c)$ for RandAR and the ELBO for DiT.
% We provide the results for IN-Val and IN-S in \fig{efficiency}.
% Runtimes are measured on a single A100 GPU. 
% We observe that RandAR shows better accuracy than DiT across different operating points while offering up to $25{\times}$ faster inference.

Here, we evaluate various generative classifiers of different model sizes in terms of their classification accuracy with respect to the total runtime measured in seconds per a single image classification task. Both RandAR and DiT classifiers are each evaluated with different numbers of NFEs in order to approximate the respective objectives, namely $p(\mathbf{x}|c)$ for RandAR and the ELBO for DiT. We provide the corresponding experimental results for both IN-Val and IN-S datasets in \fig{efficiency}. All of the runtimes are measured on a single A100 GPU. We observe that RandAR consistently shows better classification accuracy than DiT across all the different operating points, while at the same time offering up to $25{\times}$ faster inference.

\subsection{Error analysis}
\label{sect:error_anal}
% DONE

\begin{figure}[t!]
\vspace{4mm}
\centering
\includegraphics[width=0.75\linewidth]{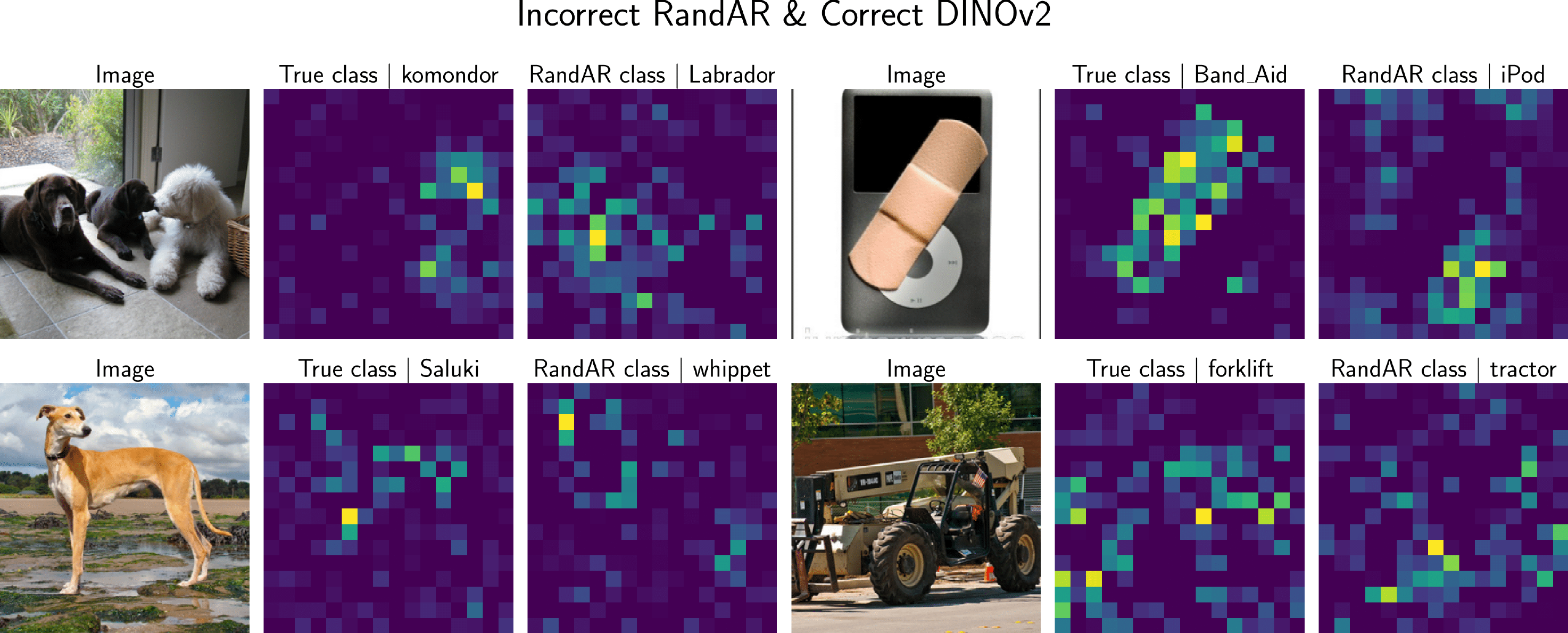}\\
\vspace{2mm}
\includegraphics[width=0.75\linewidth]{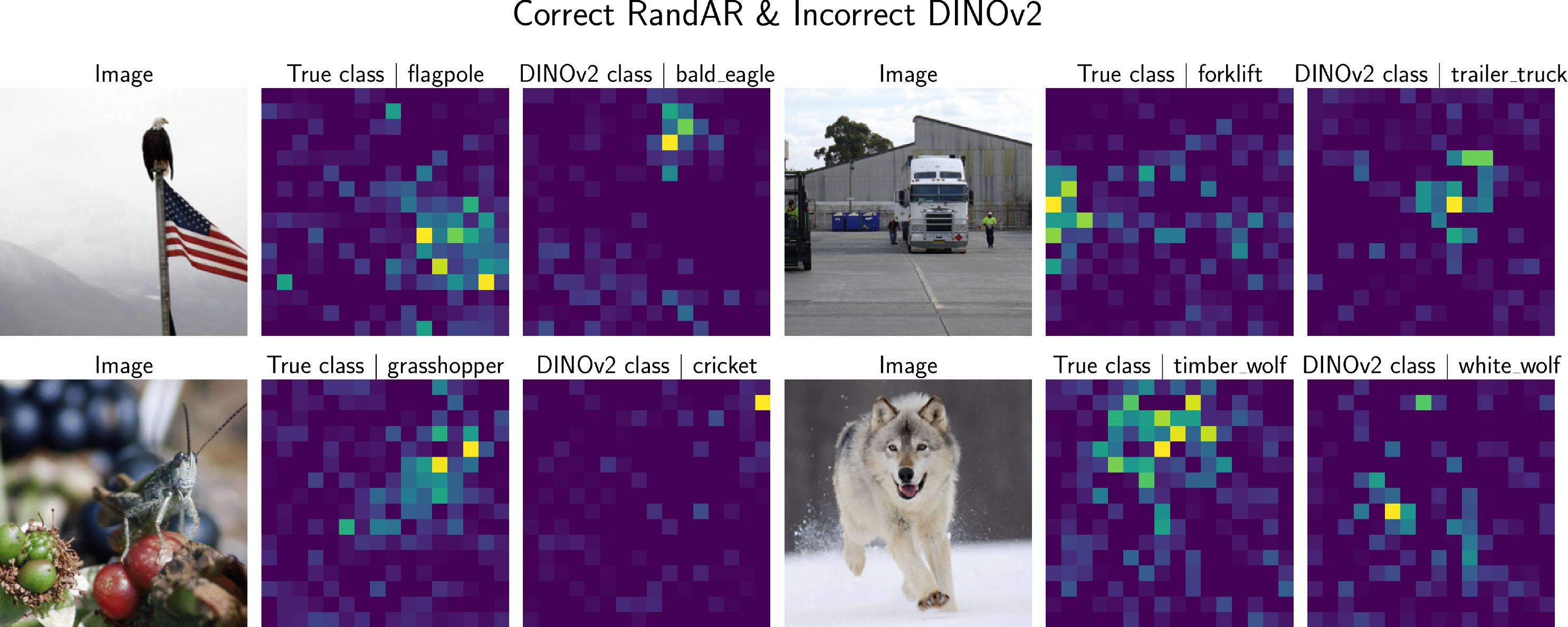}
% \vspace{-1mm}
\caption{
    \textbf{Error analysis.} 
    The examples where RandAR-XL is incorrect while DINOv2-XL is correct (Top), and vice versa (Bottom). 
    Each example is accompanied by discriminative log-likelihoods produced by RandAR for the correct and incorrect classes.
    Both models tend to fail on multi-object images or visually similar object categories.
}

\label{fig:error_anal}
\end{figure}

We analyze image examples where RandAR-XL fails while DINOv2-XL succeeds, and vice versa.
In \fig{error_anal}, we present representative examples illustrating typical failure cases of both models, along with the corresponding RandAR discriminative log-likelihoods computed for the true and predicted class labels.

We observe that both models exhibit similar error types, which are common among image classifiers and can be broadly grouped into two categories:

\begin{itemize}
    \item \textbf{Similar object classes} -- the models misclassify the objects that appear visually similar to another class, e.g., similar dog breeds.
    In such cases, the discriminative log-likelihoods are often less interpretable, as the model may assign high likelihoods to the same object.
    
    \item \textbf{Multi-object images} -- the input images contain multiple distinct objects belonging to different classes, and the model simply selects one of them as the primary prediction.
\end{itemize}

Notably, we observe that even when predicting an incorrect overall class, RandAR often still assigns high likelihood scores to the correct object when it is explicitly conditioned on the true class label.

\newpage
\begin{table}[t]
\vspace{6mm}
\begin{minipage}{0.45\linewidth}

\resizebox{\linewidth}{!}{
\begin{tabular}{l|c|c|c}
\toprule
VQ-VAE & Vocab size & Flipped tokens & rFID \\
\midrule
\multirow{4}{*}{MaskGIT} & \multirow{4}{*}{$1024$} & $0\%$ & $\mathbf{1.67}$ \\
     & & $10\%$ & $1.71$ \\
     & & $20\%$ & $1.85$ \\
     & & $40\%$ & $2.49$ \\
\midrule
\multirow{4}{*}{LlamaGen} & \multirow{4}{*}{$16384$} & $0\%$ &  $\mathbf{1.68}$ \\
      & & $18\%$ & $\mathbf{1.67}$ \\
      & & $27\%$ & $\mathbf{1.69}$ \\
      & & $42\%$ & $1.73$ \\
\bottomrule
\end{tabular}}
\vspace{1mm}
\caption{
    Reconstruction quality of the MaskGIT and LlamaGen VQ-VAE tokenizers for varying levels of latent noise corruption, which alters a portion of tokens in the original token sequences.
}
\label{tab:vqvae_comparison}
\end{minipage} \hfill
\begin{minipage}{0.52\linewidth}
\resizebox{\linewidth}{!}{
\begin{tabular}{c|c|cccc}
\toprule
Model size & Tokenizer & IN-Val & IN-R & IN-S & IN-A \\
\midrule

\multirow{4}{*}{L/16}
  & MaskGIT
  & $\mathbf{0.780}$ & $0.448$ & $0.379$ & $0.101$ \\
\cmidrule(l){2-6}
  & LlamaGen
  & $0.769$ & $\mathbf{0.469}$ & $\mathbf{0.409}$ & $\mathbf{0.146}$ \\
\cmidrule(l){2-6}
  & \makecell[l]{LlamaGen\\ + Noise aug.}
  & $\mathbf{0.780}$ & $0.463$ & $\mathbf{0.406}$ & $\mathbf{0.145}$ \\

\midrule

\multirow{3}{*}{XL/16}
  & LlamaGen
  & $\mathbf{0.812}$ & $0.512$ & $0.452$ & $\mathbf{0.241}$ \\
\cmidrule(l){2-6}
  & \makecell[l]{LlamaGen\\ + Noise aug.}
  & $\mathbf{0.813}$ & $\mathbf{0.530}$ & $\mathbf{0.459}$ & $0.233$ \\

\bottomrule
\end{tabular}

}
\vspace{1mm}
\caption{
    Comparison of the RandAR models pretrained with different tokenizers and the proposed noise augmentation in terms of \mbox{top-1} classification accuracy.
}
\label{tab:vq_vae_accuracy}
\end{minipage}
\end{table}

\subsection{The effect of VQ-VAE tokenizers}
\label{sect:vqvae_analysis}

% Though the AR models allow explicit likelihood estimation, it is calculated for the distribution induced by VQ-VAE tokenizers.
% Thus, prior to approaching main experiments, it is important to address which tokenizer to choose and how their properties affect classification performance.

% In this experiment, we follow the same model setups as in \sect{random_vs_fixed} and explore two pretrained VQ-VAE models, MaskGIT~\cite{} and LlamaGen~\cite{}, widely used by recent discrete generative models.
% These tokenizers have different codebook sizes of $1024$ in MaskGIT and $16384$ in LlamaGen.

% Though the AR models allow explicit likelihood estimation, it is calculated for the distribution induced by VQ-VAE tokenizers.
In this experiment, we evaluate the RandAR-L models for two VQ-VAE tokenizers, MaskGIT~\cite{chang2022maskgit} and LlamaGen~\cite{sun2024autoregressive}, widely used by the recent discrete generative models.
These tokenizers have different codebook sizes of $1024$ in MaskGIT and $16384$ in LlamaGen.

Prior work~\cite{yue2025understand, voronov2024switti} has noticed that tokenizers producing significantly different token sequences for almost identical images may bring additional challenges for AR model training.
Therefore, we first investigate the robustness of the MaskGIT and LlamaGen tokenizers to minor noise perturbations.

We evaluate the tokenizers in terms of their reconstruction quality while corrupting the continuous image latents (prior to quantization) with flow matching~\cite{lipman2022flow} noising process: $(1-t) z + t \epsilon$, where $\epsilon \sim \mathcal{N}(0,I)$.
We apply noising across different $t$ and track the ratio of ``flipped'' tokens in a discrete sequence and rFID~\cite{heusel2017gans} on IN-Val.
\tab{vqvae_comparison} shows that the original tokenizers provide similar reconstruction quality and thus can be fairly compared to each other.
Then, we observe that the larger codebook is resistant to ${\sim}27\%$ flipped tokens while the small one starts degrading with ${\sim}10\%$.

The results suggest applying the noise corruption technique as an augmentation during RandAR training with the LlamaGen tokenizer to make the model invariant to slight perturbations, not leading to noticeable image differences.
Meanwhile, this augmentation is not used for the MaskGIT tokenizer since only few tokens can be changed without compromising reconstruction quality. 

Finally, we compare RandAR models pretrained with different VQ-VAE tokenizers, including the one using the noise augmentation.
\tab{vq_vae_accuracy} shows that RandAR with MaskGIT provides higher in-domain accuracy while LlamaGen shows slightly better robustness across all OOD benchmarks.
Noise augmentation improves LlamaGen's in-domain accuracy while maintaining comparable OOD performance.
Therefore, in all our experiments, we use the LlamaGen tokenizer with the suggested noise augmentation.

% Noise augmentation does not show consistent improvements but we believe it still can be useful for small training datasets. 

% \newpage
\subsection{Likelihood estimation strategy ablation}

\begin{wraptable}{l}{0.44\linewidth}
\vspace{-6mm}
\resizebox{\linewidth}{!}{
\begin{tabular}{l|ccccc}
\toprule
Strategy $\diagdown$ $K$  &  $1$ & $2$ & $5$ & $10$ & $20$ \\
\midrule
 Lower bound & $0.712$ & $0.762$ & $0.795$ & $0.804$ & $0.813$ \\
 Log-likelihood & $0.712$ & $0.724$ & $0.742$ & $0.747$ & $0.760$ \\
\bottomrule
\end{tabular}}
\vspace{-2mm}
\caption{Ablation of likelihood estimation strategies for order-marginalized AR GCs on IN-val.}
\label{tab:lowerbound_vs_likelihood}
\vspace{-4mm}
\end{wraptable}

Here, we ablate the order-marginalized likelihood estimation approach, discussed in~\Cref{sect:method}.
In~\tab{lowerbound_vs_likelihood}, we compare direct log-likelihood estimation, $\log \mathop{\mathbb{E}}_{\pi} \left[ p(\mathbf{x} | \pi, c) \right]$, with its lower bound, $\mathop{\mathbb{E}}_{\pi} \left[ \log p(\mathbf{x} | \pi, c) \right]$.
We observe that the lower-bound estimate achieves significantly higher classification accuracy as $K$ increases.

\subsection{Generalization beyond RandAR}

In our work, we primarily focus on RandAR, as it is the only publicly available model that directly supports random-order generation. 
However, to demonstrate the generalization of the order-marginalized approach, we also consider another order-conditioned architecture, RAR~\cite{yu2025randomized}. 

RAR is not natively any-order, since its random permutations are gradually annealed to raster order during training. 
We therefore train RAR-B without order annealing to enable any-order sampling. 
As shown in \tab{masked_diffusion} (left), RAR-B matches RandAR-B and outperforms DiT-B, suggesting that the benefits of order marginalization are not specific to RandAR.

\subsection{Any-order AR as masked diffusion}
\label{sect:masked_diffusion}

As discussed in Section~\ref{sect:method}, any-order AR (AO-AR) models are closely related to masked diffusion (MD) models~\cite{hoogeboom2022autoregressive, shih2022training, ou2024absorbing}.
This connection allows RandAR to be evaluated in the MD manner, where tokens beyond the observed prefix are treated as masked and scored conditionally on the prefix.
Unlike AR evaluation, which estimates the full $\log p(\mathbf{x},|,\pi, c)$ in a single forward pass, MD scores token log-likelihoods only for a fixed prefix length per pass, requiring multiple forward passes to cover all terms.

First, we compare RandAR-B under both evaluation regimes and include MaskGIT~\cite{chang2022maskgit} as an open-source MD baseline. 
Since the official MaskGIT checkpoints and training code are unavailable, we evaluate the only publicly available MaskGIT-B checkpoint\footnote{https://huggingface.co/llvictorll/Maskgit-pytorch} we found. 
In~\tab{masked_diffusion} (Left), RandAR-B significantly outperforms both MD counterparts under the same compute budget ($K{=}20$).

Then, we evaluate RandAR-B under both regimes as a function of the number of evaluations $K$.
\tab{masked_diffusion} (Right) shows that the MD-based classification approaches the AR variant only for large $K$, while AR evaluation remains significantly more accurate for small $K$.

\begin{table}[h]
% \vspace{-2mm}
\centering
\begin{minipage}[h]{0.46\linewidth}
\centering\small
\begin{adjustbox}{width=\linewidth}
\setlength{\tabcolsep}{2pt}
\begin{tabular}{l|cccc}
\toprule
Model ($K{=}20$) & IN-Val & IN-R & IN-S & IN-A \\
\midrule
DiT-B ($250$ steps) & $0.654$ & $0.280$ & $0.216$ & $0.040$ \\
\midrule
MaskGIT-B & $0.605$ & $0.315$ & $0.234$ & $0.035$ \\
RandAR-B as MD & $0.662$ & $0.329$ & $0.239$ & $0.051$ \\
\midrule
RAR-B & $\mathbf{0.741}$ & $\mathbf{0.350}$ & $\mathbf{0.303}$ & $\mathbf{0.068}$ \\
RandAR-B & $0.708$ & $\mathbf{0.354}$ & $\mathbf{0.305}$ & $\mathbf{0.058}$ \\
\bottomrule
\end{tabular}
\end{adjustbox}
\end{minipage}
\hfill
\begin{minipage}[h]{0.53\linewidth}
\centering\small
\begin{adjustbox}{width=\linewidth}
\setlength{\tabcolsep}{2pt}
\begin{tabular}{l|cccccc}
\toprule
Model $\diagdown$ $K$ & $1$ & $2$ & $5$ & $10$ & $20$ & $60$ \\
\midrule
RandAR-B as MD & 0.348 & 0.422 & 0.550 & 0.593 & 0.662 & 0.701\\
RandAR-B & $\mathbf{0.586}$ & $\mathbf{0.640}$ & $\mathbf{0.677}$ & $\mathbf{0.692}$ & $\mathbf{0.708}$ & $\mathbf{0.711}$\\
\bottomrule
\end{tabular}
\end{adjustbox}
\end{minipage}

\vspace{2mm}
\caption{
\textbf{Left:} B-scale comparisons on ImageNet benchmarks for $K{=}20$ including MaskGIT, RAR, and RandAR evaluated as masked diffusion.
\textbf{Right:} ImageNet-Val accuracy as a function of the number of evaluations $K$ for RandAR-B scored as either an any-order AR model or masked diffusion.
}
\label{tab:masked_diffusion}
% \vspace{-4mm}
\end{table}

% \vspace{-1mm}
\subsection{Real-world distribution shifts}
Finally, we evaluate RandAR in terms of robustness to real-world distribution shifts.
RandAR is pretrained on each dataset for B/16 and L/16 model settings using the LlamaGen tokenizer with the proposed noise augmentation.
\begin{table*}[h]
\vspace{-2mm}
\centering
\begin{adjustbox}{width=0.85\textwidth}
\begin{tabular}{c|l|cc|cc|cc}
\toprule
{Model size} & Method & \multicolumn{2}{c|}{Camelyon17} & \multicolumn{2}{c|}{CelebA} & \multicolumn{2}{c}{FMoW}\\
& & ID & OOD & ID & ID WG & ID & OOD WG\\
\midrule
\multirow{6}{*}{B/16} 
 & ERM~\cite{vapnik2013nature}      & $0.946$ & $0.581$ & $\mathbf{0.939}$ & $0.342$ & $0.310$ & $0.137$ \\
 & RWY~\cite{sagawa2019distributionally}      & $0.945$ & $0.578$ & $0.923$ & $0.426$ & $0.291$ & $0.139$ \\
 & $\sigma{-}\text{stitching}$~\cite{puli2023don} & $0.938$ & $0.667$ & $0.926$ & $0.461$ & $-$ & $-$\\
 & DiT~\cite{peebles2023scalable}      & $\mathbf{0.992}$ & $0.586$ & $0.899$ & $\mathbf{0.601}$ & $\mathbf{0.571}$ & $\mathbf{0.281}$ \\
 & RandAR raster~\cite{pang2025randar} & $\mathbf{0.990}$ & $0.647$ & $0.897$ & $0.572$ & $0.423$ & $0.205$ \\
 & RandAR~\cite{pang2025randar} \textbf{(Ours)}   & $\mathbf{0.992}$ & $\mathbf{0.783}$ & $0.908$ & $\mathbf{0.600}$ & $0.563$ & $\mathbf{0.276}$ \\
\midrule
\multirow{6}{*}{L/16}
 & ERM~\cite{vapnik2013nature}   & $0.954$ & $0.538$ & $\mathbf{0.939}$ & $0.359$ & $0.304$ & $0.140$ \\
 & RWY~\cite{sagawa2019distributionally}   & $0.952$ & $0.543$ & $0.914$ & $0.453$ & $0.279$ & $0.130$ \\
 & $\sigma{-}\text{stitching}$~\cite{puli2023don} & $0.942$ & $0.677$ & $0.922$ & $0.419$ & $-$ & $-$\\
 & DiT~\cite{peebles2023scalable}   & $\mathbf{0.995}$ & $0.525$ & $0.904$ & $0.596$ & $\mathbf{0.584}$ & $\mathbf{0.274}$ \\
 & RandAR raster~\cite{pang2025randar} & $\mathbf{0.996}$ & $0.501$ & $0.893$ & $0.594$ & $0.346$ & $0.189$ \\
 & RandAR~\cite{pang2025randar} \textbf{(Ours)} & $\mathbf{0.996}$ & $\mathbf{0.763}$ & $0.920$ & $\mathbf{0.639}$ & $0.567$ & $\mathbf{0.273}$ \\
\bottomrule
\end{tabular}
\end{adjustbox}
\vspace{2mm}
\caption{Classification performance comparison on the datasets with subpopulation or distribution shifts.}
\label{tab:wilds_results}
\end{table*}

\vspace{-6mm}
\paragraph{Datasets.}
These experiments are conducted on the WILDS benchmark~\cite{koh2021wilds}, which contains a collection of datasets for evaluating robustness to distribution shifts.
Specifically, we consider the CelebA~\cite{liu2015deep}, FMoW~\cite{christie2018functional}, and Camelyon~\cite{bandi2018detection} datasets that represent a diverse range of shift types and visual domains.
CelebA consists of natural face images and exhibits a subpopulation shift, and we therefore report worst-group (WG) accuracy.
FMoW includes satellite images and involves both subpopulation shifts and a domain shift, for which we report OOD worst-group accuracy.
Camelyon contains tissue patches from different hospitals.

\paragraph{Baselines.}
Among discriminative baselines, we consider ERM~\cite{vapnik2013nature}, RWY~\cite{sagawa2019distributionally, idrissi2022simple} and ERM with the $\sigma{-}$stitching loss~\cite{puli2023don}.
As a backbone, we use ViT~\cite{dosovitskiy2020image} and train the models from scratch following~\cite{li2024generative}.
We train $3$ models with different random seeds and average the results.
As a generative classifier baseline, we consider DiT~\cite{peebles2023scalable}, training the B/16 and L/16 variants at an image resolution of $256{\times}256$.

\paragraph{Model selection.}
We follow the setup from~\cite{li2024generative}.
Specifically, checkpoints are selected based on the class-balanced accuracy on in-domain validation set.

\paragraph{Results.}
\tab{wilds_results} provides the results for both ID and OOD test sets.
On Camelyon17, RandAR outperforms the supervised baselines and matches DiT on the ID set, while substantially surpassing all baselines on the OOD set.
On CelebA, ERM achieves higher in-domain accuracy but performs poorly on the OOD set. 
RandAR exceeds DiT on the ID set and outperforms all models in WG accuracy in most cases.
On FMoW, generative classifiers largely outperform discriminative methods on the ID set, although DiT shows slightly higher accuracy than RandAR.
We attribute this to the relatively small training set of  ${\sim}77K$, where DiT may be more sample-efficient than RandAR due to stronger noise regularization.
In terms of OOD WG accuracy, RandAR and DiT perform similarly and both largely surpass ERM and RWY.

In total, RandAR demonstrates high in-domain accuracy and strong robustness to subpopulation and distribution shifts, integrally outperforming DiT.

\section{Conclusion}

In this work, we investigate recent visual AR models as GCs and introduce a framework that leverages AR models generating images in random token orders. 
We show that order-marginalized AR classifiers significantly outperform diffusion-based and prior AR counterparts, achieving state-of-the-art performance in generative image classification.
These findings suggest that enabling AR models to represent images using different token orders opens new perspectives, and we look forward to further development and scaling of any-order AR models to better understand the limits of order-marginalized AR classifiers.

A promising direction for future research is to leverage advances in self-supervised learning, which have already shown strong potential in raster-order AR models~\cite{yue2025understand} and diffusion models~\cite{yu2024representation}. 
Another interesting update would be to equip RandAR classifiers with image-adaptive token-order predictors, inspired by recent results in diffusion-based classifiers~\cite{wang2025noise}.

Moreover, since one of the key limitations of GCs is their significant computational cost, particularly for the problems with a large number of classes, it would be valuable to explore whether these classifiers can be effectively distilled into discriminative models. 
Such an approach could potentially combine the best of both worlds: the high inference efficiency of discriminative models and the strong classification performance of generative classifiers.

% Also, it would be interesting to investigate such AR classifiers across different modalities, e.g., tabular data or time-series, where foundational and powerful pretrained backbones are missing.

% \input{tables/lecture_jit_tmp}
\bibliographystyle{splncs04}
\bibliography{main}

\newpage
\clearpage
\setcounter{page}{1}
\appendix

% \section*{Supplementary Material}

\section{Model settings}
\label{app:setup}

\paragraph{ImageNet}
\begin{itemize}
    \item \textbf{RandAR}:
    In our main experiments, we train RandAR from scratch using the LlamaGen tokenizer, following the setup of the official RandAR implementation. 
    We use an AdamW optimizer with a learning-rate schedule with $40{,}000$ warm-up steps to a peak of $6{\times}10^{-4}$, followed by cosine decay over the next $460{,}000$ iterations. 
    Training is conducted with a batch size of $512$ for RandAR-L and $256$ for RandAR-XL, with all other hyperparameters retained from the default setup. 
    We also apply random crop and latent noise augmentations during training.
    
    \item \textbf{DiT}: The models are trained from scratch following the setup in the DiT repository\footnote{https://github.com/facebookresearch/DiT}.
    Training is performed for $400$ epochs using a AdamW optimizer, with a constant learning rate of $1{\times}10^{-4}$ and a batch size of $256$. 
    % For the DiT-XL/2 variant, we use the pretrained weights from the official repository.
    Similar to RandAR, we apply random crop augmentation during training that noticeably improves its classification performance.
    Note that our DiT-XL model shows slightly better classification performance than the released one.
    
    \item \textbf{DINOv2}: Since the released models are pretrained on ImageNet-21K, we use the official code and configuration to train both L/16 and XL/16 models on ImageNet1K for images of $256{\times}256$ resolution. 
    We train the models for $200$ epochs using a batch size of $512$.

    \item \textbf{SiT:} We evaluate the publicly available SiT-XL/2 model\footnote{https://github.com/willisma/SiT}. 
    We also evaluate the representation-aligned (REPA) SiT-XL/2 checkpoint\footnote{https://github.com/sihyun-yu/REPA} using the same evaluation protocol.
    
    \item \textbf{ViT:} We evaluate the public ImageNet pretrained ViT-L/16 model\footnote{https://github.com/lucidrains/vit-pytorch}. % and its official script for evaluation.

    \item \textbf{VAR:} 
    We use the publicly available VAR-L and VAR-XL checkpoints\footnote{https://github.com/FoundationVision/VAR}.
    
    \item \textbf{A-VARC+:} We follow the official code\footnote{https://github.com/Yi-Chung-Chen/A-VARC}.
    \item \textbf{DAT:} We use the pretrained weights from the official DAT repository\footnote{https://github.com/xuwangyin/DAT}.
\end{itemize}

\paragraph{WILDS}
\begin{itemize}

    \item \textbf{RandAR}:
    The model is trained using the LlamaGen tokenizer with noise augmentation, as discussed in Section 6.5.
    We also apply random crop augmentation to reduce overfitting.
    For training, we use a learning rate of $1{\times}10^{-4}$ for the CelebA and FMOW datasets, and $2{\times}10^{-4}$ for Camelyon17.
    The batch size is set to $256$ for CelebA and Camelyon17, and $128$ for FMOW. 
    All models are trained for $150{,}000$ iterations per dataset.

    \item \textbf{DiT}: The models are trained with a learning rate of $1{\times}10^{-4}$ for all datasets. 
    Similarly, to reduce overfitting, we apply random crop augmentation to the input images.
    The batch size is $256$ for CelebA and Camelyon17, and $128$ for FMOW. 
    Each model is trained for $300{,}000$ iterations.

    \item \textbf{ERM/RWY}: 
    We train ERM and RWY using the official WILDS codebase\footnote{https://github.com/p-lambda/wilds}. 
    Each model is trained until convergence, and the best checkpoint is selected based on validation balanced accuracy.

\end{itemize}

\begin{figure}[h]
\begin{minipage}{0.42\linewidth}
\vspace{1mm}
\centering
\resizebox{0.95\linewidth}{!}{
\begin{tabular}{l|ccc|c}
\toprule
Model & NFE & T & N & IN-Val \\
\midrule
 \multirow{5}{*}{DiT-XL}
   & $100$ & $50$ & $2$ & $0.748$ \\
   & $200$ & $50$ & $4$ & $0.749$ \\
   \cmidrule(lr){2-5}
   & $100$ & $100$ & $1$ & $0.765$ \\
   & $200$ & $100$ & $2$ & $0.761$ \\
   \cmidrule(lr){2-5}
   & $200$ & $200$ & $1$ & $0.768$ \\
 \midrule
 % RandAR-XL & 2 & - & 2 & $0.752$ \\
 \multirow{2}{*}{RandAR-XL} & 4 & - & - & $0.781$ \\
  & 20 & - & - & $\mathbf{0.815}$ \\
\bottomrule
\end{tabular}}
\captionof{table}{Comparison to diffusion classifiers with averaging over $N$ noise samples per timestep and $T$ timesteps. 
Noise averaging per timestep does not provide extra gains. 
Also, it is consistently better to use more timesteps for ELBO estimation under the same budget.}
\label{tab:averaging}
\end{minipage}
\hfill
\begin{minipage}{0.56\linewidth}
\includegraphics[width=\linewidth]{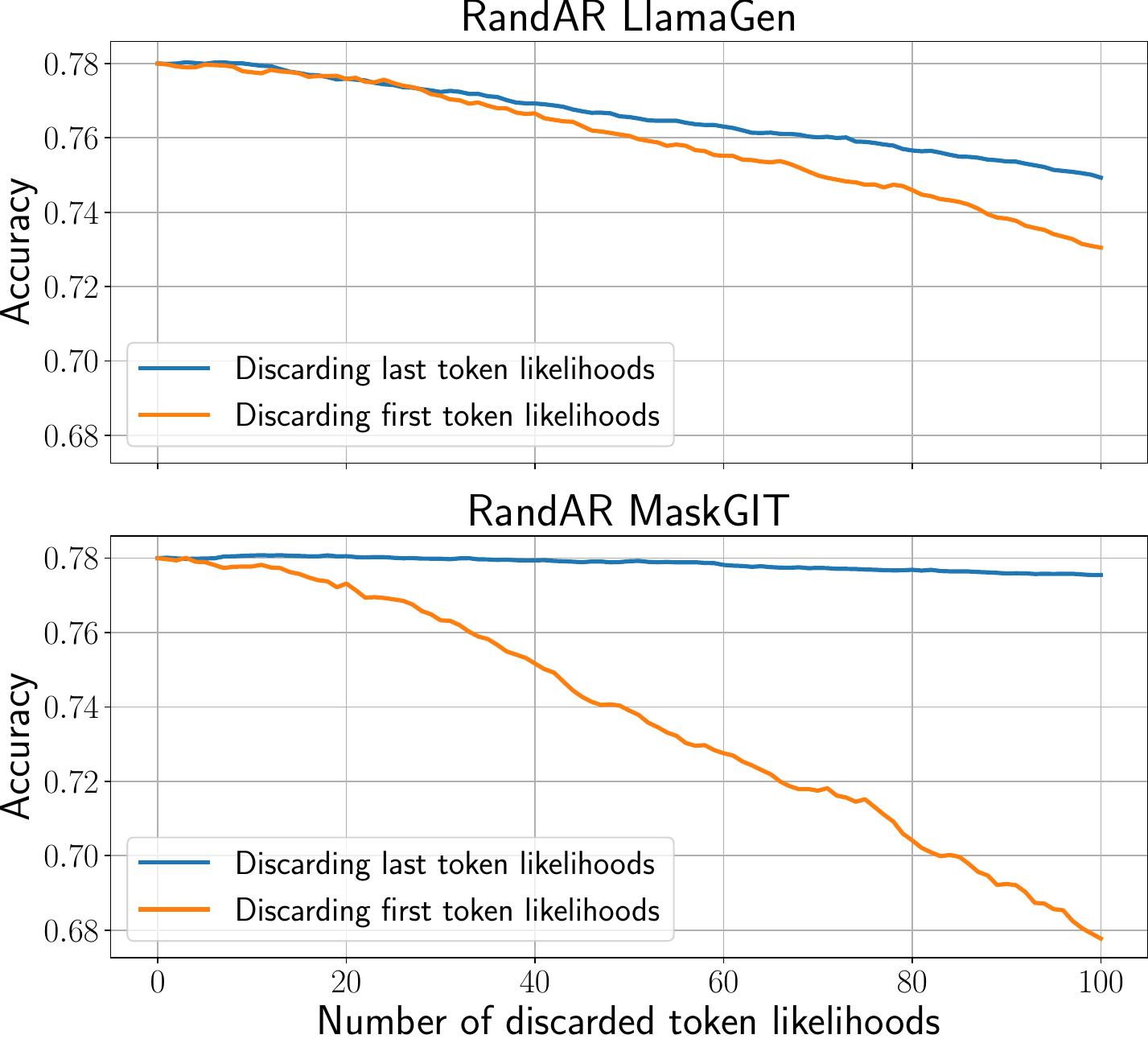}
\caption{
    Classification performance when the first and last predicted token likelihoods are discarded in the overall log-likelihood estimation.
}
\label{fig:dropped_tokens}
\end{minipage}
\end{figure}

% \todo{
% We noticed that random crop augmentation during training significantly improve classification capabilities, while preserving generative performance (see ~\Cref{tab:crop_effect}). 
% Therefore, we apply augmentations for both RandAR and DiT models.}

% \section{Uncertainty}
% In~\fig{uncertainty_xl}, we provide additional uncertainty estimation results for RandAR-XL and DINOv2-XL models.
% Consistently to the results in~\fig{uncertainty}, RandAR-XL also demonstrates more interpretable uncertainty estimates than DINOv2-XL.

\section{Analysis of per-token likelihood importance}
\label{app:token_importance}
In this section, we extend the analysis in~\Cref{fig:per_token_accuracy_by_prefix} and investigate if all token log-likelihoods contribute to the overall performance.
\fig{dropped_tokens} presents the classification performance on $10K$ images from IN-Val for $K{=}20$ when the first and last $k$ predicted log-likelihoods are discarded for the overall log-likelihood estimation.
We evaluate the RandAR-L models pretrained with the LlamaGen (Top) and MaskGIT (Bottom) tokenizers.
The results indicate that removing log probabilities of tokens from the end of a sequence causes less accuracy degradation than removing them from the beginning. 
Interestingly, with the MaskGIT tokenizer, optimal performance can be achieved by using the first ${\sim}200$ out of $256$ token predictions. 
In contrast, achieving the best results with the LlamaGen tokenizer requires utilizing the log probabilities of all tokens in a sequence.

\section{Averaging predictions for Diffusion Classifiers}
\label{app:average_dm}

Diffusion classifiers allow averaging across timesteps and noise samples to estimate ELBO (\eq{elbo}). 
By default, we average across timesteps and use different noise samples at each timestep, as proposed in prior work~\cite{li2023your}. 
Additionally, we may further average the DM predictions across multiple noise samples at each timestep.

~\tab{averaging} shows the DM results for different numbers of timesteps ($T$) and noise samples per timestep ($N$).
We observe that averaging over multiple noise samples at each timestep does not provide noticeable gains, while averaging across timesteps remains more effective under the same computational budget.

\section{Equivalence of any-order AR and masked diffusion}
\label{app:ao_ar_md}

Here, we show that the order-marginalized any-order AR (AO-AR) objective used in~\Cref{eq:randar_likelihood} coincides, in expectation, with the masked diffusion (MD) evidence lower bound~\cite{hoogeboom2022autoregressive, shih2022training, ou2024absorbing}.
Let $N$ denote the sequence length, $\pi$ a permutation of token indices, and $M$ a randomly masked subset of tokens with observed complement $\bar{M}$.
Starting from the order-averaged log-likelihood and rearranging the expectations over orders and positions yields:
\begin{equation}
\label{eq:ao_ar_md_elbo}
\begin{aligned}
& \mathbb{E}_{\pi} \log p(\mathbf{x} \mid \pi, c) =  \mathbb{E}_{\pi} \frac{1}{N} \sum_{n=1}^{N}
\log p\!\left(\mathbf{x}_{\pi(n)} \mid \mathbf{x}_{\pi(<n)}, c\right) = \\
&= \mathbb{E}_{\pi} \mathbb{E}_n \log p\!\left(\mathbf{x}_{\pi(n)} \mid \mathbf{x}_{\pi(<n)}, c\right) = \\
&= \mathbb{E}_n \mathbb{E}_{\pi} \log p\!\left(\mathbf{x}_{\pi(n)} \mid \mathbf{x}_{\pi(<n)}, c\right) = \\
&= \mathbb{E}_n \mathbb{E}_{\pi} \frac{1}{N-n+1} \sum_{k \in \pi(\ge n)} \log p\!\left(\mathbf{x}_{\pi(k)} \mid \mathbf{x}_{\pi(<t)}, c\right) = \\
&= \mathbb{E}_M \frac{1}{|M|}
\sum_{i \in M}
\log p_\theta\!\left(\mathbf{x}_i \mid \mathbf{x}_{\bar{M}}, c\right) = ELBO_{\text{MD}}(\mathbf{x}, c).
\end{aligned}
\end{equation}

Consequently, MaskGIT~\cite{chang2022maskgit}, which is trained with $ELBO_{\text{MD}}$, can be used directly as a generative classifier. 
Similarly, the RandAR model can be scored in the MD manner by observing only the unmasked tokens $\mathbf{x}_{\bar{M}}$ at each step.
Nevertheless, we emphasize that despite the equivalence in expectation, estimating this expectation in either the AR or MD manner leads to different practical effectiveness.
We empirically compare both regimes in~\Cref{sect:masked_diffusion}.

\section{Loss weighting for Diffusion Classifiers}
\label{app:loss_weighting}

We ablate different loss weighting approaches for DiT and SiT to compare our AR approach with stronger baselines. 
We formulate the weighting schemes in the $\epsilon$-pred form, $w(t)\cdot\lVert \epsilon_\theta - \epsilon \rVert$. 
Note that the $\mathbf{v}$- and $\mathbf{x}$-pred forms are equivalent to $\epsilon$-pred up to the choice of $w(t)$.
We consider several popular weighting schemes: constant, ELBO~\cite{kingma2021variational}, SNR$^{-1}$, Min-SNR-5~\cite{Hang_2023_ICCV} and FM-OT~\cite{lipman2022flow}, summarized here~\cite{kingma2023understanding}.
We evaluate the officially released DiT-XL and SiT-XL checkpoints.
\tab{weighting_dm} shows that the constant weighting, used in our main experiments, consistently outperforms alternatives for both DiT and SiT.

\begin{table*}[t]
\centering
\begin{tabular}{l|ccccc}
\toprule
Model & Constant & ELBO & SNR$^{-1}$ & Min-SNR-5 & FM-OT\\
\midrule
 DiT-XL & $\mathbf{0.751}$ & $0.663$ & $0.126$ & 0.742 & 0.179 \\
 SiT-XL & $\mathbf{0.685}$ & $0.632$ & $0.373$ & 0.682 & $0.576$ \\
\bottomrule
\end{tabular}
\vspace{1mm}
\caption{Weighting strategy ablation for diffusion classifiers on IN-val.}
\label{tab:weighting_dm}
\end{table*}

\begin{table*}[t]
\centering
\begin{tabular}{l|cccccc}
\toprule
 Method & IN-Val & IN-R & IN-S & IN-C Gauss & IN-C JPEG & IN-A \\
\midrule
 ResNet-18 & $0.706$ & $0.331$ & $0.212$ & $0.195$ & $0.504$ & $0.012$ \\
 ResNet-34   & $0.731$ & $0.361$ & $0.243$ & $0.301$ & $0.554$ & $0.019$ \\
 ResNet-50   & $0.777$ & $0.362$ & $0.260$ & $0.300$ & $0.588$ & $0.000$ \\
 ResNet-101   & $0.777$ & $0.393$ & $0.278$ & $0.391$ & $0.618$ & $0.047$ \\
 ResNet-152      & $0.783$ & $0.413$ & $0.308$ & $0.443$ & $0.646$ & $0.061$ \\
% \midrule
%  A-VARC d16 ~\cite{chen2025your} (10 augs)  & $0.683$ & $0.233$ & $0.157$ & $0.155$ & $0.377$ & $0.082$ \\
%  A-VARC d20 ~\cite{chen2025your} (10 augs)  & $0.661$ & $0.199$ & $0.138$ & $0.077$ & $0.315$ & $0.090$ \\
\midrule
 RandAR L/16 ~\cite{pang2025randar} \textbf{(Ours)}& $\textbf{0.780}$ & $\textbf{0.463}$ & $\textbf{0.406}$ & $\textbf{0.589}$ & $\textbf{0.687}$ & $\textbf{0.145}$ \\
 
 RandAR XL/16 ~\cite{pang2025randar} \textbf{(Ours)} & $\textbf{0.813}$ & $\textbf{0.530}$ & $\textbf{0.459}$ & $\textbf{0.652}$ & $\textbf{0.751}$ & $\textbf{0.233}$ \\
\bottomrule
\end{tabular}
\vspace{1mm}
\caption{Comparison to the ResNet models in top-1 accuracy on the ImageNet benchmarks for different model sizes.}
\vspace{-7mm}
\label{tab:ResNets}
\end{table*}

\section{Comparison to additional baselines}
\label{app:additional_evals}

\paragraph{Comparison to ResNet classifiers.}
Following~\cite{li2023your,chen2025your}, we also include comparisons with convolutional-based classifiers~\cite{he2016residual} in~\tab{ResNets}.
We observe that our RandAR models (both L and XL) consistently outperform ResNets.

\begin{table*}[t]
\centering
\begin{tabular}{l|c|cccc}
\toprule
Method & IN-Val & IN-R & IN-S & IN-A \\
\midrule
 SiT-XL  & $0.685$ & $0.249$ & $0.219$ & $0.102$\\
 SiT-XL + REG~\cite{wu2025representation}   & $0.451$ & $0.103$ & $0.060$ & $0.045$\\
 SiT-XL + REPA~\cite{yu2024representation}   & $0.727$ & $0.290$ & $0.261$ & $0.168$\\
 SiT-XL + iREPA~\cite{singh2025matters}   & $0.708$ & $0.325$ & $0.293$ & $0.131$\\
 \midrule
 RandAR-XL & $\mathbf{0.813}$ & $\mathbf{0.530}$ & $\mathbf{0.459}$ & $\mathbf{0.233}$\\
\bottomrule
\end{tabular}
\vspace{1mm}
\caption{Comparison of RandAR with SSL aligned DMs.}
\label{tab:SSL}
\end{table*}

\paragraph{Extended comparison to representation-aligned DMs.}
We also compare our approach to different REPA variants~\cite{yu2024representation,singh2025matters,wu2025representation} applied to DMs. 
The results in \tab{SSL} show that the original REPA~\cite{yu2024representation} achieves higher accuracy than the other REPA variants, but still underperforms our AR-based approach.

\section{Non image-centric and multi-object datasets}
While our main experiments focus on ImageNet-style image-centric benchmarks, we additionally evaluate whether our approach extends to more diverse settings.
Specifically, we consider \textbf{ObjectNet} \cite{barbu2019objectnet}, which contains non-image-centric object views, and \textbf{COCO} \cite{lin2014microsoft}, which contains multiple objects per image. 
For COCO, we use an COCO-to-ImageNet label mapping\footnote{https://github.com/howardyclo/ImageNet2COCO} and report $K$-recall@$K$, where $K$ is the number of ImageNet-mapped objects in the image.

\tab{SSL} shows that RandAR significantly outperforms DiT on both datasets, while DINOv2 remains a stronger competitor.

\begin{table}[h]
\vspace{-2mm}
\centering
\begin{tabular}{c|l|c|c}
\toprule
Model size & Model & ObjectNet & COCO \\
\midrule
\multirow{3}{*}{L}
& DINOv2 & $0.439$ & $0.411$ \\
& DiT & $0.308$ & $0.368$ \\
& RandAR & $0.339$ & $0.378$ \\
\midrule
\multirow{3}{*}{XL}
& DINOv2 & $0.455$ & $0.418$ \\
& DiT & $0.321$ & $0.369$ \\
& RandAR & $0.397$ & $0.404$ \\
\bottomrule
\end{tabular}
\vspace{2mm}
\caption{Generalization of RandAR to non image-centric datasets.}
\label{tab:SSL}
\end{table}

\section{Generation quality vs.\ classification accuracy}
Here, we examine whether generation quality strongly correlates with classification accuracy in generative models. 
We observe that better generation quality does not necessarily imply higher accuracy. 
We hypothesize that this is because FID reflects both class understanding and perceptual fidelity, with the latter becoming more dominant at low FID values. 
A similar observation was reported in A-VARC+~\cite{chen2025your}.

\begin{table}[h]
\vspace{-3mm}
\centering
\setlength{\tabcolsep}{2pt}
\begin{tabular}{l|cc|l|cc}
\toprule
Model & FID$\downarrow$ & Acc.$\uparrow$ & Model & FID$\downarrow$ & Acc.$\uparrow$\\
\midrule
LlamaGen-L & $3.80$ & $0.640$ & LlamaGen-XL & $2.62$ & $0.559$ \\ 
DiT-L & $2.97$ & $0.769$ & DiT-XL & $2.27$ & $0.767$ \\
RandAR-L & $2.90$ & $0.780$ & RandAR-XL & $2.34$ & $0.813$ \\
\bottomrule
\end{tabular}
\vspace{2mm}
\caption{FID vs.\ ImageNet-Val accuracy for different generative classifiers.
Lower FID does not necessarily imply higher classification accuracy.}
\end{table}

\newpage

\section{Comparison to few-step diffusion}
\label{app:few-step}
This work compares our AR-based method against standard multi-step diffusion models (e.g., $250$-step DiT/SiT).
However, given the large variety of high-quality few-step diffusion-based models~\cite{geng2026mean, song2023consistency, yin2024one, sauer2024adversarial}, a convincing argument about practical speed advantages requires a comparison against a state-of-the-art few-step diffusion baseline. 
Therefore, we compare against \textbf{MeanFlow} (MF)~\cite{geng2026mean}.

Note that few-step generation does not directly imply that these models allow fast and accurate likelihood estimation.
Therefore, to examine whether recent few-step diffusion models can serve as GCs, we evaluate MeanFlow under several classification strategies:
\begin{itemize}
    \item \textbf{Diffusion:} treating MF as a standard diffusion model by setting $t=r$ and computing the diffusion ELBO as for the diffusion baselines.

    \item \textbf{Few-step:} exploiting the few-step sampling property, we classify by the label that minimizes the error between one-step $\mathbf{x}$ predictions at different timesteps $t$ and the corresponding data sample.  

    \item \textbf{Training objective (JVP):} classifying each image by the label that minimizes the MF training objective, including the JVP term.
\end{itemize}

Across the best parameterizations and schedules we found, all variants largely underperform DiT at comparable runtime.
Other few-step alternatives we tried, including improved MeanFlow, performed worse. 

\begin{figure}[h]
    \centering
    % \vspace{-3mm}
    \includegraphics[width=0.95\linewidth]{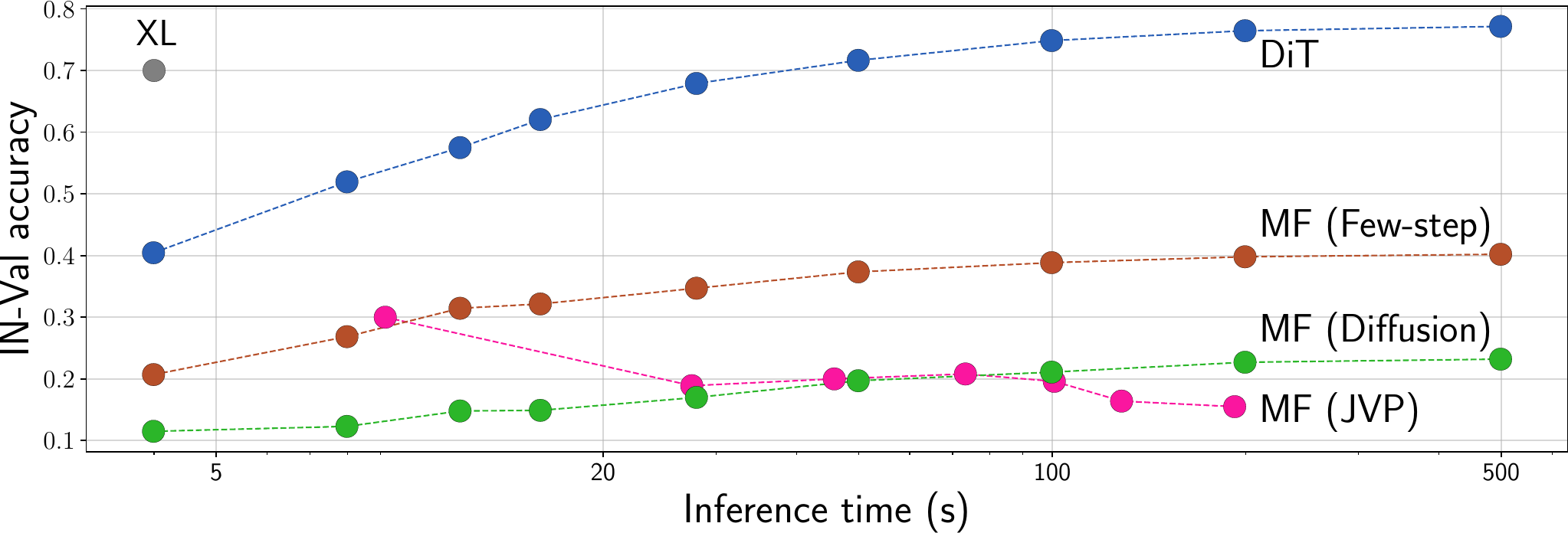}
    \label{fig:meanflow_eval}
    \caption{Accuracy-runtime trade-off on ImageNet-Val for DiT and MeanFlow-based generative classifiers.}
\end{figure}

\newpage
\section{More examples of per-token log-likelihoods}
\label{app:per_token_likelihoods}
\begin{figure*}[b!]
\centering
\includegraphics[width=0.8\linewidth]{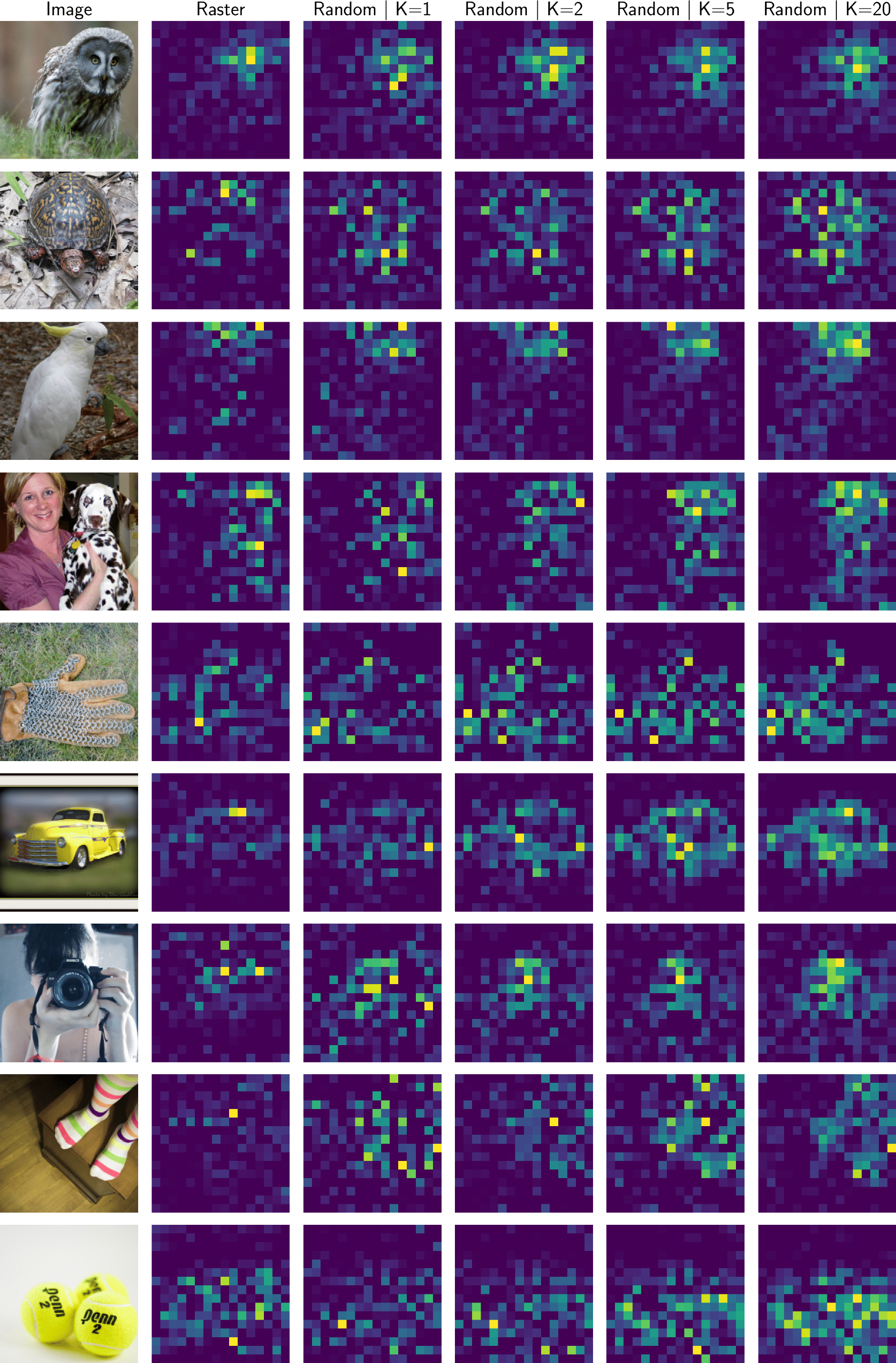}
\caption{
    % DONE
    Per-token ``discriminative'' log-likelihoods, computed as $clip[\log p(\mathbf{x} | c_{true}) - \log p(\mathbf{x} | c_{false}), 0]$, across different orders and $K$ values.
    $c_{true}$ denotes the correct class; $c_{false}$ refers to a randomly selected incorrect one.
    Order-marginalized log-likelihood estimates ($K>1$) capture class-specific attributes more accurately.
}
\label{fig:per_token_likelihoods_ext}
\end{figure*}

In \fig{per_token_likelihoods_ext}, we provide more visual examples of discriminative per-token log-likelihoods obtained with RandAR for different $K$ settings.

\end{document}